\documentclass{article}

\usepackage{microtype}
\usepackage{graphicx}
\usepackage{subfigure}
\usepackage{booktabs} 
\usepackage{lipsum}
\usepackage{siunitx}

\usepackage{hyperref}

\newcommand{\R}{\mathbb R}

\usepackage[accepted]{icml2024}

\usepackage{amsmath}
\usepackage{amssymb}
\usepackage{mathtools}
\usepackage{amsthm}

\usepackage[capitalize,noabbrev]{cleveref}

\theoremstyle{plain}
\newtheorem{theorem}{Theorem}[section]

\newtheorem{lemma}{Lemma}

\theoremstyle{definition}

\theoremstyle{remark}

\usepackage[textsize=tiny]{todonotes}

\icmltitlerunning{Partially Stochastic Infinitely Deep Bayesian Neural Networks}

\begin{document}

\twocolumn[
\icmltitle{Partially Stochastic Infinitely Deep Bayesian Neural Networks}



\icmlsetsymbol{equal}{*}

\begin{icmlauthorlist}
\icmlauthor{Sergio Calvo-Ordoñez}{equal,omi,ox}
\icmlauthor{Matthieu Meunier}{equal,ox}
\icmlauthor{Francesco Piatti}{equal,imp}
\icmlauthor{Yuantao Shi}{equal,omi,ox}
\end{icmlauthorlist}

\icmlaffiliation{omi}{Oxford-Man Institute of Quantitative Finance, University of Oxford}
\icmlaffiliation{ox}{Mathematical Institute, University of Oxford}
\icmlaffiliation{imp}{Department of Mathematics, Imperial College London}

\icmlcorrespondingauthor{Sergio Calvo-Ordoñez}{sergio.calvoordonez@maths.ox.ac.uk}

\icmlkeywords{Machine Learning, ICML}

\vskip 0.3in
]



\printAffiliationsAndNotice{\icmlEqualContribution} 

\begin{abstract}
In this paper, we present Partially Stochastic Infinitely Deep Bayesian Neural Networks, a novel family of architectures that integrates partial stochasticity into the framework of infinitely deep neural networks. Our new class of architectures is designed to improve the computational efficiency of existing architectures at training and inference time. To do this, we leverage the advantages of partial stochasticity in the infinite-depth limit which include the benefits of full stochasticity e.g. robustness, uncertainty quantification, and memory efficiency, whilst improving their limitations around computational complexity. We present a variety of architectural configurations, offering flexibility in network design including different methods for weight partition. We also provide mathematical guarantees on the expressivity of our models by establishing that our network family qualifies as Universal Conditional Distribution Approximators. Lastly, empirical evaluations across multiple tasks show that our proposed architectures achieve better downstream task performance and uncertainty quantification than their counterparts while being significantly more efficient. The code can be found at \url{https://github.com/Sergio20f/part_stoch_inf_deep}
\end{abstract}
\vspace{-0.15in}
\section{Introduction}

By conceptualising neural networks as a composition of infinitely many residual layers, Neural ODEs \cite{Neural_ODEs} transform the output of a network into the solution of an ODE \cite{Haber_2017}. This approach of infinite-depth parametrisation has two primary advantages: it permits a nuanced balance between computational cost and precision via adaptive computation tailored to specific error tolerances, and it notably diminishes memory requirements during training. The latter is achieved by using the adjoint method which allows reverse-time reconstruction of states, facilitating the use of such models in time series and diffusion applications \citep{calvoordonez2023missing}.

Previous work \citep{xu2022infinitely} proposed a framework for implementing Bayesian Neural Networks (BNNs) in this infinite-depth limit using Neural SDEs \citep{liu2019neural}. These models offer a probabilistic interpretation of continuous-time neural networks by placing a prior distribution over the network's weights and updating them through Bayesian inference steps. This involves inferring distributions over parameters rather than relying on point estimates (\citealt{article-mackay, Neal1995BayesianLF}). This approach introduces several key benefits like accounting for model uncertainty (\citealt{bergna2023graph, vanamersfoort2020uncertainty, pmlr-v124-ustyuzhaninov20a, he2023recombiner}), which is key in high-stakes environments such as financial forecasting (\citealt{arroyo2024deep, moreno2022deepvol}) or healthcare (\citealt{vanderschueren2023accounting}).

However, for both continuous and discrete BNNs, the process of estimating the posterior distribution of network weights typically involves computationally intensive techniques like Markov Chain Monte Carlo or variational inference methods, both of which can be resource-heavy and time-consuming (\citealt{Neal1995BayesianLF, blundell2015weight}). These methods often require a large number of iterations or complex optimisation procedures to accurately approximate the posterior distribution. Additionally, the need to store and process multiple samples for each weight during training and inference further escalates the computational load.

To overcome these challenges, previous research has explored partially stochastic networks as a potential solution in finite-depth settings (\citealt{daxberger2022bayesian, kristiadi2020bayesian, li2024training}). These networks restrict stochasticity to a subset of the model’s parameters, denoted as $\Theta = \Theta_S \cup \Theta_D$, and the learning process involves estimating a distribution for $\Theta_S$ while obtaining point estimates for $\Theta_D$. These methods aim to achieve greater computational efficiency while retaining the benefits of Bayesian approaches. Initially, the main idea of these architectures was to approximate fully stochastic networks with a trade-off between computational efficiency and accuracy. However, Sharma et al. \yrcite{do_BNN_need_fully_stochastic} provide both theoretical and empirical evidence to demonstrate that partially stochastic networks can achieve comparable, if not superior, performance to their fully stochastic counterparts. While Sharma et al. \yrcite{do_BNN_need_fully_stochastic} leaves the selection of non-stochastic components in BNNs largely unexplored, works focusing on Variational Bayesian Last Layers like Harrison et al. \yrcite{harrison2024variational}, address this issue. As there is no trivial way to split the weights into subsets of deterministic and stochastic weights in the infinite-depth limit, previous research has only explored finite-depth regimes. However, studies such as Farquhar et al. \yrcite{gal_liberty_or_depth} hint towards the viability of extending this approach to an infinite-depth configuration.

Building on these insights, our work introduces Partially Stochastic Infinitely Deep Bayesian Neural Networks (PSDE-BNNs). This class of architectures combines the flexibility of neural differential equations with the efficient uncertainty quantification of partially stochastic networks.
By selectively introducing noise into the weights process, our model achieves an enhanced representation of uncertainty and superior performance in downstream tasks compared to fully stochastic counterparts. Additionally, it significantly reduces the computational overhead typically associated with infinitely deep BNNs. Our contributions to the field are summarised as follows:

\begin{itemize}
    \item We introduce the concept of partial stochasticity within continuous-time neural models, expanding the existing framework of Bayesian Neural Networks.

    \item We provide a detailed mathematical framework to incorporate partial stochasticity into the evolution of the weights of infinitely deep BNNs. 

    \item We prove mathematically under which specific conditions infinitely deep BNNs fail to be Universal Conditional Distribution Approximators (UCDAs). Conversely, we demonstrate that Partially Stochastic Infinitely Deep BNNs, under different conditions, do qualify as UCDAs.

    \item We demonstrate that our class of models outperforms fully stochastic networks in downstream tasks and uncertainty quantification, showing that full stochasticity may not be necessary. We further conduct ablation studies on key PSDE-BNN hyperparameters.
\end{itemize}

\section{Background and Preliminaries}

We build upon existing infinite-depth networks, as well as concepts of partial stochasticity in discrete networks.

\paragraph{Neural Ordinary Differential Equations} A Neural ODE \cite{Neural_ODEs} consists of a continuous-time hidden state model, whose rate of change is given by a neural network. Given an input \(x\), a Neural ODE can be written as
\[
\frac{\mathrm d h_t}{\mathrm d t} = f_
\theta(t,h_t), \quad h_0 = x,
\]
where \(f_
\theta\) is a Lipschitz function defined by a neural network with parameters \(\theta\).
To evaluate the Neural ODE at an input \(x\), we use an ODE solver, and backpropagation is computed using the adjoint sensitivity method \cite{scalable_gradients_for_SDE}.

\paragraph{Partial Stochasticity in Bayesian Neural Networks} In Bayesian Neural Networks \cite{lampinen2001bayesian,titterington2004bayesian}, parameters are treated as random variables, with the aim to infer their posterior distribution \(p(w|\mathcal{D})\) given a dataset \(\mathcal{D}\) and prior \(p(w)\). Approximate inference involves finding an approximate posterior, often through a parametrisation \(q_\theta(w)\). Specifically, variational inference methods intend to maximise the Evidence Lower Bound (ELBO), or equivalently, minimise the Kullback-Leibler (KL) divergence between \(q_\theta(w)\) and \(p(w|\mathcal{D})\). The gradient of this objective can be computed using Monte-Carlo samples via Stochastic Variational Inference \cite{hoffman13-svi}. In addition, previous studies suggest that full stochasticity is unnecessary for effective posterior approximation in finite-depth BNNs \citep{do_BNN_need_fully_stochastic}. Extending these findings, we show that partial stochasticity preserves the expressivity of BNNs in the infinite-depth regime.

\paragraph{SDEs as Approximate Posteriors}
Stochastic differential equations introduce Brownian motion as a source of continuous noise in the dynamics of a time-dependent state \(X_t\) \cite{oksendal2013stochastic}. Formally, a \(d\)-dimensional stochastic differential equation driven by a \(m\)-dimensional Brownian motion \(B_t\) on a time interval \([0,T]\) is often written as 
\[
\mathrm d X_t = \mu(t,X_t) \mathrm d t + \sigma(t, X_t) \mathrm d B_t,
\]
where \(\mu: [0,T] \times \mathbb R^d \rightarrow \mathbb R^d \) (the drift), \(\sigma: [0,T]\times \mathbb R^d \rightarrow \mathbb R^{d \times m}\) (the volatility or diffusion matrix) are Lipschitz functions in the state, and satisfy a linear growth condition. Such conditions ensure the existence and uniqueness of a strong solution, given a square-integrable initial condition \(X_0\). SDEs are a natural way of embedding stochasticity in a continuous-time framework, hence they are fit for modeling infinitely deep networks in a Bayesian setting. For these networks, Xu et al. \yrcite{xu2022infinitely} propose to model the prior on weights \((w_t)_{t \in [0,1]}\) as an SDE 
\begin{equation}
\label{eq:background-prior-sde}
    \mathrm d w_t = f_p (t,w_t) \mathrm d t + g_p(t,w_t) \mathrm d B_t,
\end{equation}
and parametrize the approximate posterior using a Neural SDE, where the drift \(f_q\) is a neural network with parameters $\theta$
\begin{equation}
    \label{eq:background-posterior-sde}
    \mathrm d w_t = f_q(t,w_t;\theta) \mathrm d t + g_p(t,w_t) \mathrm d B_t.
\end{equation}
Variational inference on SDEs was introduced by Archambeau et al. \yrcite{archambeau2007variational}, and previous work has relied on other techniques to learn the coefficients of SDEs, such as the Karhunen-Loève expansion of Brownian motion \cite{ghosh2022differentiable}.
 The parametrization of the weights given by \eqref{eq:background-posterior-sde} requires using KL divergence on path spaces. Minimizing this divergence between path measures induced by diffusion processes is a well-studied mathematical problem \cite{cattiaux1994minimization} which has been recently leveraged in the context of learning the Schrödinger Bridge Problem \cite{vargas2023bayesian} and regularizing generative models \cite{richter2024improved, vargas2024transport}. We use the following infinite-dimensional version of the KL divergence \cite{tzen2019neural,scalable_gradients_for_SDE}
\begin{align*}
D_{\mathrm{KL}}\left(\mu_q \| \mu_p\right) & =\mathbb{E}_{q_\theta(w)}\left[\int_0^1 \frac{1}{2}\|u_\theta(t, w_t)\|_2^2 \mathrm{~d} t\right] \text { where } \\
u_\theta(t, w_t) & =g_p\left(t,w_t\right)^{-1}\left[f_q\left(t,w_t; \theta\right)-f_p\left(t,w_t\right)\right],
\end{align*}
with \(\mu_p,\mu_q\) being the prior and posterior measures on the path space \(\mathcal{P} \subset \mathbb R ^{[0,1]}\) induced by \eqref{eq:background-prior-sde} and \eqref{eq:background-posterior-sde} respectively. We show that when only a fraction of the weights are stochastic, such approximate posteriors yield UCDAs.


\section{Method}
Standard discrete-depth residual networks \cite{he2016deep} can be defined as a composition of layers of the form
\begin{equation}
    h_{t+\delta} = h_t + \delta f(h_t;w_t), \quad t=1,\dots,T,
\end{equation}
where \(t\) is the layer index, \(h_t\in\mathbb{R}^{d_h}\) represents a vector of hidden unit activations at layer $t$, the input is \(h_0=x\), \(w_t\) are the parameters for layer t, and $\delta=1$. Taking \(\delta = 1 / T\) and letting \(T \rightarrow \infty\) yields a Neural ODE \(\mathrm d h_t = f_h(t,h_t;w_t)\mathrm d t\), where $f_h$ is a neural network with parameters $w_t$. These parameters can also be modelled by a neural network (more precisely, a hypernetwork), which in the infinite-depth limit becomes itself a neural differential equation \(\mathrm d w_t = f_q(t,w_t; \theta) \mathrm d t\) (or, in the Bayesian setting, a Neural SDE). Therefore, $f_q$ is a neural network parametrised by $\theta$ which models the continuous evolution of the parameters of $f_h$. Finally, in this framework, the only trainable parameters are the vector $\theta$ and $w_{0}$. While we use an SDE formalism similar to generative diffusion models, our model is \textit{discriminative}. It ultimately learns the conditional distribution $p(y \mid x)$ via continuous latent variables.

We now present a detailed mathematical formulation for the class of Partially Stochastic Infinitely Deep BNNs.

\subsection{Vertical Separation of the Weights}

\begin{figure}[t]  
    \centering
    \includegraphics[width=0.45\textwidth]{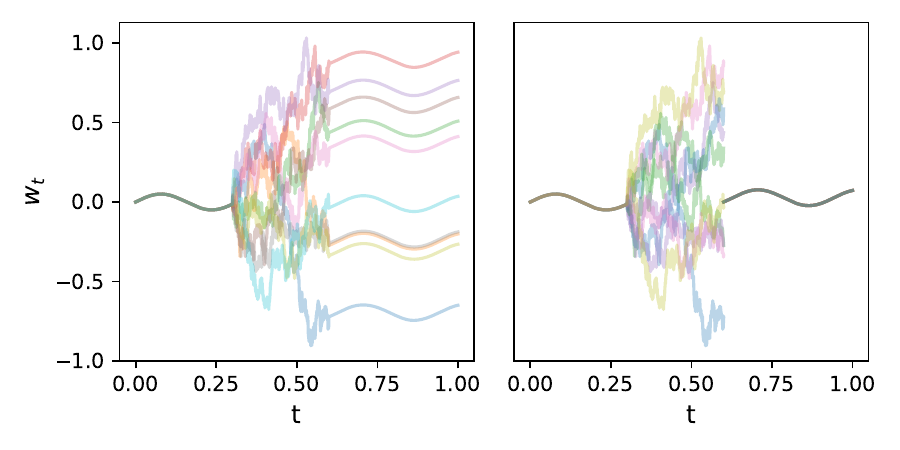}
    \caption{\textbf{Vertical Cut:} Sample paths of $w_t$ when (left) not fixing $w_{t_2}$ vs. (right) fixing $w_{t_2}$. If we do not fix $w_{t_2}$, $w_t$ will be random in the interval $(0.6, 1)$. Here, $f_q = \cos(20t)$, $g_p = 0$ for $t \notin (0.3, 0.6)$, and $f_q = 0$, $g_p = 1$ for $t \in (0.3, 0.6)$.}
    \vspace{-0.15in}
    \label{fig: sample-paths}
\end{figure}

First, we consider a vertical separation of the weights in the neural network $f_q$ (Figure \ref{fig: sample-paths}). Let \(w_t \in \mathbb R^{d_w}\) represent the weights of layer \(t \in [0,1]\), and \(t_1 < t_2\) be cutoff points of the interval where the weights evolve stochastically, then consider an infinite-depth network architecture such that:
\begin{itemize}
    \item For \(t \notin (t_1,t_2)\), the dynamics of \(w_t\) are deterministic, i.e
    \begin{equation}
    \label{eq:weights_deterministic}
        \mathrm d w_t = f_q(t,w_t; \theta) \mathrm d t,
    \end{equation}
    where \(f_q: [0,1]\times \mathbb{R}^{d_w}\to\mathbb{R}^{d_w}\) is a neural network parametrised by \(\theta \in \R^{d_{\theta}}\).
    \item For \(t \in (t_1,t_2)\), we treat \(w_t\) as a random vector, and perform Bayesian inference. To that end, we define a prior as the solution of the SDE
    \begin{equation}
    \label{eq:weights_prior}
        \mathrm d w_t = f_{p}(t,w_t)\mathrm d t + g_{p}(t,w_t) \mathrm d B_t,
    \end{equation}
    as well as an approximate posterior 
    \begin{equation}
    \label{eq:weights_posterior}
        \mathrm d w_t = f_q(t,w_t; \theta)\mathrm d t + g_{p}(t,w_t) \mathrm d B_t,
    \end{equation}
    where \(B_t\) is a standard \(d_w\)-Brownian motion, and \(f_p,g_p\) are deterministic functions fixed \textit{a priori}. 
\end{itemize}
Notice that the diffusion of the approximate posterior is fixed to match the prior diffusion. This is to ensure that the KL divergence between the prior and posterior measures on paths $\mu_p$ and $\mu_q$ respectively, is finite. This can be formulated in the following lemma (see Archambeau et al., \citeyear{archambeau2007variational} for a proof).

\begin{lemma}\label{lemma:KL-diffusion}
    Assume the prior and posterior of $w_t$ are given as 
    \begin{equation}
        \mathrm d w_t = f_{p}(t,w_t)\mathrm d t + \sigma_{p}(t,w_t) \mathrm d B_t,
    \end{equation}
    \begin{equation}
    \label{eq:weights_posterior_KL_proof}
        \mathrm d w_t = f_{q}(t,w_t)\mathrm d t + \sigma_{q}(t,w_t) \mathrm d B_t.
    \end{equation}    
    Assume that \(\sigma_p,\sigma_q\) are continuous. If there exists \(t \in (0,1)\) such that $\sigma_{q}(t,w_t)\neq \sigma_{p}(t,w_t)$ with non-zero probability, then $D_{\mathrm{KL}}\left(\mu_{q} \| \mu_{p}\right)=\infty$.
\end{lemma}

Over the whole time interval \([0,1]\), the hidden state dynamics \(h_t \in \R^{d_h}\) is given by 
\begin{equation}
\label{eq:hidden_state_dynamics}
    \mathrm d h_t = f_h(t,h_t;w_t) \mathrm dt,
\end{equation}
where \(f_h(t,h_t;w_t)\) is again a neural network whose parameters are $w_t$.

In our model, the points $t_1$ and $t_2$ play a key role in dictating the behaviour of the neural network parameters $w_t$. Specifically, $w_{t_1}$ denotes the initial condition of the SDE, which, if $t_1 > 0$, also serves as the terminal value of the ODE governing the weight evolution in the interval $[0, t_1)$. Conversely, $w_{t_2}$ is the sampled state at the SDE's endpoint. When $t_2<1$, deciding the initial value for the subsequent ODE spanning $[t_2, 1]$ becomes crucial. The first approach is to continue with the same vector $w_{t_2}$, allowing for a continuous transition into the ODE. Alternatively, the initial value for this ODE could be fixed \textit{a priori} or set as a learnable parameter. This choice significantly influences the network's parameter trajectory: using the same $w_{t_2}$ introduces an element of randomness beyond $t_2$, as illustrated in Figure \ref{fig: sample-paths} (left), where $w_t$ is non-deterministic in the interval $(0.6, 1)$. On the other hand, fixing $w_{t_2}$ \textit{a priori}, as shown in Figure \ref{fig: sample-paths} (right), ensures that $w_t$ remains deterministic throughout the interval $(0.6, 1)$.



\begin{figure*}[t]
    \centering
    \includegraphics[width=0.8\textwidth]{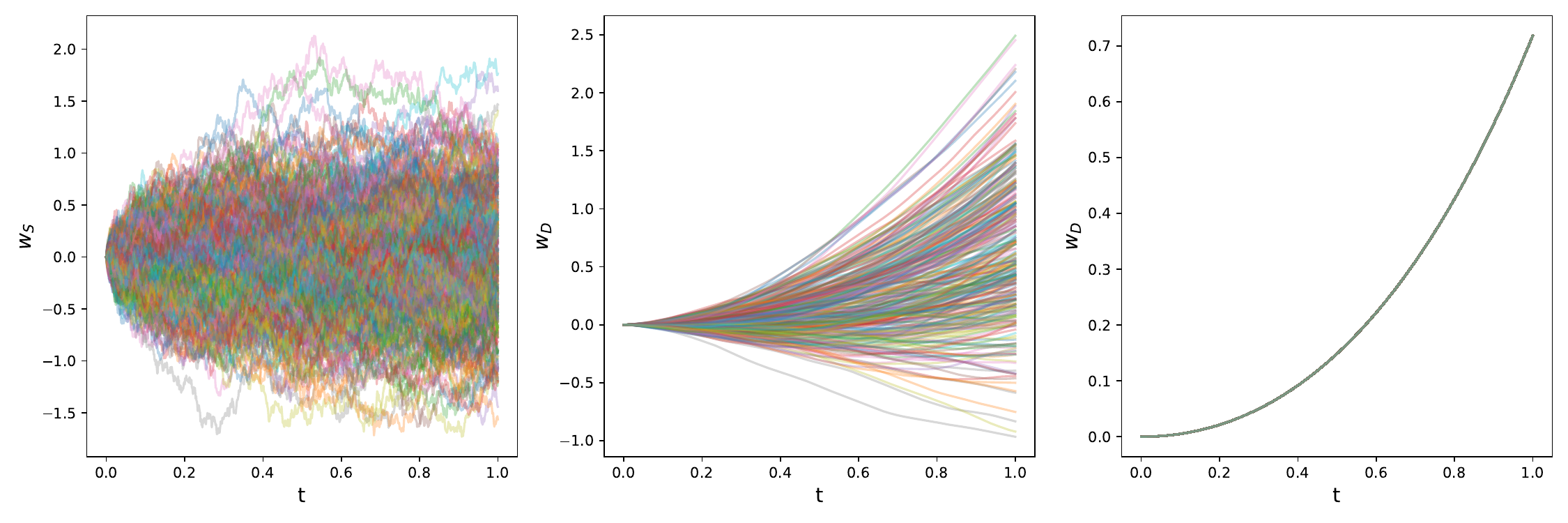}
    \vspace{-0.15in}
    \caption{\textbf{Horizontal Cut:} Sample paths of \(w_S\) (left) and $w_D$ when not separating $f_q$ (middle) vs. separating $f_q$ (right). If we do not separate $f_q$, then $w_{D}$ will also be random. Here, when not separating $f_q$, we take $f_q=[-w_S,t+w_D+w_S]$ and $g_p=[1,0]$, when separating $f_q$, we take $f_q=[-w_S,t+w_D]$ and $g_p=[1,0]$.}
    \label{fig: horizontal separation}
    \vspace{-0.1in}
\end{figure*}

In conclusion, the dynamics of the whole system can then be written as a single SDE in \(\R^{d_h+d_w}\):
\begin{align}
    \label{eq:system_sde}
    \mathrm d \begin{bmatrix} h_t \\ w_t \end{bmatrix} = 
    \begin{bmatrix}
        f_h(t,h_t;w_t) \\ f_q(t,w_t; \theta)
    \end{bmatrix} \mathrm d t +
    \begin{bmatrix}
        0 \\
        g_p(t,w_t)
    \end{bmatrix} \mathrm d B_t,
\end{align}
where we impose \(g_{p}(t,w_t) = 0\) for \(t \notin (t_1,t_2)\).
For the initial conditions, we impose \(h_0 = a(x)\), where \(x\) is the input data point and \(a : \R^{d_x} \rightarrow \R^{d_h}\) corresponds to a pre-processing step. The initial weight \(w_0 \in \R^{d_w}\) is a trainable parameter. Note that \eqref{eq:system_sde} is a slight abuse of notation, since $w_t$ might jump at time $t_2$.



\subsection{Horizontal Separation of the weights}

In \eqref{eq:system_sde}, $g_p(t,w_t):\mathbb{R}\times\mathbb{R}^{d_w}\rightarrow\mathbb{R}^{d_w}\times\mathbb{R}^{d_w}$ denotes the diffusion matrix of $w_t$. Similar to Sharma et al. \yrcite{do_BNN_need_fully_stochastic}, where parameters in a single hidden layer are separated into deterministic and stochastic groups, we can perform this parameter separation in the same layer if we set the diffusion matrix $g_p(t,w_t)$ to zero in our SDE setting, e.g.
$$
g_p(t,w_t)=
    \begin{bmatrix}
        \sigma I_{m_{1}}\quad \mathbf{0}_{m_1\times m_2} \\
        \mathbf{0}_{m_2\times m_1}\quad \mathbf{0}_{m_{2}}\\
    \end{bmatrix}.
$$
Refer to Figure \ref{fig: horizontal separation} for an intuitive understanding where $m_1=m_2=1$. In this configuration, only the first $m_1$ elements of $w_t$ interact with the Brownian motion $B_t$, and are thus random, while the remaining $m_2$ elements are deterministic. We term this division a ``horizontal separation". It is important to highlight that, within this horizontal separation framework, the function $f_q(t, w_t; \theta)$ must also be split, and therefore distinctly formulated for the stochastic ($S$) and deterministic ($D$) weights:
$$
    f_q \ = \ 
    \begin{bmatrix}
        f_{\theta_S}(t,w_t^S; \theta_S) \\
        f_{\theta_D}(t,w_t^D; \theta_D)\\
    \end{bmatrix},
$$
where $w_t^S$ represents the subset of $w_t$ corresponding to the indices in $S$, this allows us to reformulate \eqref{eq:system_sde} as follows:
\begin{align}
    \label{eq:system_sde_horizontal}
    \mathrm d \begin{bmatrix} h_t \\ w_t^D\\w_t^S \end{bmatrix} = 
    \begin{bmatrix}
        f_h(t,h_t;w_t^S, w_t^D) \\ f_{\theta_{D}}(t,w_t^D;\theta_D)\\f_{\theta_{S}}(t,w_t^S;\theta_S)
    \end{bmatrix} \mathrm d t +
    \begin{bmatrix}
        0 \\ 0 \\
        g_p(t,w_t^S)
    \end{bmatrix} \mathrm d B_t.
\end{align}
The rationale behind separating $f_{q}$ stems from the necessity of maintaining the deterministic nature of $w_t^D$ in this setting. Otherwise, if $f_{\theta_D}$ were to incorporate $w_t^S$ as an input, it would introduce stochastic elements into $f_{\theta_D}$, and consequently, $w_t^D$ would no longer remain deterministic.

For a better understanding, we refer to Figure \ref{fig: horizontal separation}, which illustrates the concept in a 2D setting for $w_t$. In the middle panel, we do not separate $f_{q}$, resulting in a random $w_t^D$, which is not desirable. In the right panel, we separate $f_{q}$, which gives a fixed $w_t^D$.\\
 

\vspace{-0.2in}
\subsection{Training the Network}

To evaluate our network given an input \(x\), we integrate \eqref{eq:system_sde}
using both a SDE and ODE solver to obtain a distribution for \(h_1\). We then use \(h_1\) to compute the likelihood of the output \(y\).
Our trainable parameters are \(\Theta = (w_0,\theta) \in \R^{d_w} \times \R^{d_{\theta}}\) (or \(\Theta = (w_0,w_{t_2},\theta) \in \R^{d_w} \times \R^{d_w}\times\R^{d_{\theta}}\) if $t_2<1$ and we set $w_{t_2}$ as a learnable parameter).
We perform approximate inference on \(w_t\), as done by Xu et al. \yrcite{xu2022infinitely}. Thus, we maximise the ELBO: \begin{equation}
\label{eq:elbo}
        \mathcal L^V(\Theta) = \mathbb E _{q_\theta} \left[ \log p (\mathcal D | w) -  \kappa \int_{t_1}^{t_2}||u_\theta(t,w_t)||^2 \mathrm d t \right]
\end{equation}
    where $ u_\theta(t, w_t)= g_p(t,w_t)^{-1}[f_p(t,w_t) - f_{q}(t,w_t; \theta)]$.
    
In \eqref{eq:elbo}, the expectation is taken with respect to the approximate posterior distribution on the whole path \((w_t)_{0 \leq t \leq 1}\). Note that $\kappa$ is a coefficient that determines the weight of the KL divergence term in the ELBO, and that \eqref{eq:elbo} corresponds to a model with vertical separation of the weights. We use the stochastic variational inference scheme of Li et al. \yrcite{scalable_gradients_for_SDE} to compute the gradient of the loss with respect to \(\Theta\). For the horizontal cut \eqref{eq:system_sde_horizontal}, the ELBO takes the form:
\begin{equation}
\nonumber
        \mathcal L ^H(\Theta) = \mathbb E _{q_\theta} \left[ \log p (\mathcal D | w) -  \kappa \int_{t_1}^{t_2}||u_\theta^S(t,w_t^S)||^2 \mathrm d t \right],
\end{equation}
    with $ u_\theta^S(t, w_t^S)= g_p(t,w_t^S)^{-1}[f_p(t,w_t^S)- f_{\theta_S}(t,w_t^S; \theta_S)]$.
As for the prior process, we use the Ornstein–Uhlenbeck (OU) process as in Xu et al. \yrcite{xu2022infinitely}. 


\section{Expressivity Guarantees}
In this section, we investigate the conditions under which our Partially Stochastic Infinitely Deep Bayesian Neural Networks qualify as Universal Conditional Distribution Approximators (UCDAs). 
\subsection{Constrained Infinitely Deep Bayesian Neural Networks are not Universal Approximators}
\label{subsec:constrainedSDEBNN_not_universal}
Dupont et al. \yrcite{dupont2019augmented} proved that a $n$-dimensional Neural ODE:
$$
\left\{\begin{array}{l}
\frac{\mathrm{d} \mathbf{v}(t)}{\mathrm{d} t}=\mathbf{f}(\mathbf{v}(t), t) \\
\mathbf{v}(0)=\mathbf{x},
\end{array}\right.
$$
with globally Lipschitz continuous $\mathbf{f}:\mathbb{R}^n\times\mathbb{R}\rightarrow\mathbb{R}^n$ and a linear output layer $\mathcal{L}:\mathbb{R}^n\rightarrow\mathbb{R}$ cannot approximate the following function $z:\mathbb{R}^n\rightarrow\mathbb{R}$
\begin{align}\label{g_non_approx}
 \begin{cases}z(\mathbf{x})=-1 & \text { if }\|\mathbf{x}\| \leq r_1 \\
z(\mathbf{x})=1 & \text { if } r_2 \leq\|\mathbf{x}\| \leq r_3,\end{cases}   
\end{align}

where $r_1<r_2<r_3$. Define the flow of the ODE given the initial value $\mathbf{x}$ as $\phi_t(\mathbf{x})=\mathbf{v}(t)$. Dupont et al. \yrcite{dupont2019augmented} mainly proved that the feature mapping $\phi_1(\mathbf{x}):\mathbb{R}^n\rightarrow\mathbb{R}^n$ is a homeomorphism and therefore it is impossible for the linear layer $\mathcal{L}$ to separate $\phi_1(\{\|\mathbf{x}\| \leq r_1\})$ and $\phi_1(\{r_2 \leq\|\mathbf{x}\| \leq r_3\})$. Similar arguments apply to our setting which we define by \eqref{eq:system_sde} with $d_x=d_h$ and $a(x)=x$ and we formalise it into the theorem below:
\begin{theorem}
    \label{theo:non-approximate}
    Let $(\Omega, \mathcal{F}, \mathbb{P})$ denote a probability space. All random variables and stochastic processes
considered are defined in that space. Consider a Linear output layer $\mathcal{L}:\mathbb{R}^{d_x}\rightarrow\mathbb{R}$ and the dynamics
    \begin{align}
    \label{eq:system_sde_non_approx}
    \mathrm d \begin{bmatrix} h_t \\ w_t \end{bmatrix} = 
    \begin{bmatrix}
        f_h(t,h_t;w_t) \\ f_q(t,w_t; \theta)
    \end{bmatrix} \mathrm d t +
    \begin{bmatrix}
        0 \\
        g_p(t,w_t)
    \end{bmatrix} \mathrm d B_t,
\end{align} with $h_0=x,w_0=0$
, where $f_h:\mathbb{R}\times\mathbb{R}^{d_x}\times\mathbb{R}^{d_w}\rightarrow\mathbb{R}^{d_x}\times\mathbb{R}^{d_w}$ is globally Lipschitz continuous in $h_t$ and continuous in $t$ and $w$, $f_q:\mathbb{R}\times\mathbb{R}^{d_w}\rightarrow\mathbb{R}^{d_w}$ is continuous in $t$ and $w$ and $g_p:\mathbb{R}\times\mathbb{R}^{d_w}\rightarrow :\mathbb{R}^{d_w}\times\mathbb{R}^{d_w}$ is Lipschitz continuous in $w_t$, continuous in $t$ and satisfies a linear growth condition. Define the flow of $h_t$ given the initial value $x$ as $\psi_t(\omega,x)=h_t$, where $\omega\in\Omega$ denotes the sample. The final output $\mathcal{L}(\psi_1(\omega,x))$ cannot represent the function $z$ defined in  \eqref{g_non_approx} in the sense of supreme norm. That is:
$$
\mathbb{P}\left(\sup_{x\in A}|\mathcal{L}(\psi_1(\omega,x))-z(x)|>1/2\right)=1,
$$
where $A=\{\|{x}\| \leq r_1,x\in\mathbb{R}^{d_x}\}\cup\{r_2 \leq\|{x}\| \leq r_3,x\in\mathbb{R}^{d_x}\}$ is the domain of $z(x)$.
\end{theorem}
The detailed proof is in Appendix \ref{Appendix:proof}.

\subsection{Partially Stochastic Infinitely Deep Bayesian Neural Networks are UCDAs}

We investigate the expressivity of our model, that is, we investigate whether it is possible to approximate any conditional distribution with a posterior as defined in our PSDE-BNN architectures. Theorem \ref{thm:classical-PSBNN} hints to get a positive result, especially because we benefit from an ``infinite-dimensional" source of noise through the Brownian Motion \(B_t\). However, since the hidden state \(h_t\) is modelled with an ODE (for a fixed weight trajectory \((w_t)_{t \in [0,1]}\)), we need to augment the input space to unlock the model's approximation power, as done by Dupont et al. \yrcite{dupont2019augmented} with the so-called Augmented Neural ODEs. Our theorem leverages four main results.

First, we assume that, for $m \in \mathbb N$ large enough, there exists a continuous \textit{generator function} for the conditional distribution $Y|X$, i.e, a function $\tilde{f}: \mathcal X \times \R^m \rightarrow \mathcal Y$ such that for any $\eta \sim \mathcal N (0,I_m)$ independent of $X$, $(X,\tilde{f}(X,\eta))$ is distributed like $(X,Y)$. This assumption is motivated by the Noise Outsourcing lemma \cite{kallenberg_foundations_2010} and we state it in Appendix \ref{Appendix:proof}. 
Second, we use the results on the approximation capabilities of Neural ODEs, first proven by Zhang et al. \yrcite{zhang20h} - Lemmas \ref{lemma:homeo-approx-net} and \ref{lemma:node-approx-net} in Appendix \ref{Appendix:proof}. Finally, we use the fact that Partially Stochastic finite-depth Bayesian Neural Networks are UCDAs \cite{do_BNN_need_fully_stochastic}, as stated in Appendix \ref{Appendix:proof}. We now present our main result:

\begin{theorem}
\label{thm:main-theorem}
    Let \(X,Y\) be random variables taking values in \(\mathcal X \subset \mathbb R^{d_x}, \mathcal Y \subset \mathbb R^{d_y}\) respectively. Assume \(\mathcal X\) is compact. Assume further that there exists a continuous generator function \(\tilde{f}: \mathbb R^m \times \mathcal X \rightarrow \mathcal Y\) for the conditional distribution \(Y | X\) for some \(m \geq d_y\). Consider a PSDE-BNN model given by \eqref{eq:system_sde}, with $d_h, d_w$ large enough. Then, for all \(\varepsilon > 0\), there exist \(A \in \mathcal F,a,f_h,f_q,g_p, \theta\), a linear operator \(\mathcal L\) and a random variable \(\eta \sim \mathcal{N}(0,I_m)\) independent of \(X\) such that 
    \begin{align*}
        (i)& \: \: \mathbb P (A) \geq 1 - \varepsilon, \\
    (ii)& \: \: \forall \omega \in A, x\in\mathcal{X}, \: ||\mathcal{L}(\psi_1(\omega, x)) - \tilde{f}(\eta(\omega), x)|| \leq \varepsilon.
    \end{align*}
    In other words, there exists a PSDE-BNN model that approximates $Y|X$ with high probability.
\end{theorem}
The detailed proof is in Appendix \ref{Appendix:proof}. At a high level, we construct a suitable PSDE-BNN architecture which, for $t_2 < 1$, allows $w_{t_2}$ to be either fixed \textit{a priori}, a learnable parameter, or the solution to the SDE in \eqref{eq:weights_posterior} at time $t_2$. 
For the architecture of \(f_q(t,w_t;\theta)\), we introduce the following structure:
\begin{itemize}
    \item We partition the whole time interval $[0,1]$ into three distinct decision regions, namely $R_1,R_2$ and $R_3$. The specifics of these regions are elaborated upon in the proof in Appendix \ref{Appendix:proof}.
    \item For each region, we employ a network architecture capable of approximating any continuous function. Such architectures include fixed-width deep ReLU networks \cite{yarotsky17} and fixed-depth wide networks with non-polynomial activations \cite{leshno93}.  The parameters for each region \(\theta_1,\theta_2\) and \(\theta_3\) are distinct but collectively form the parameter vector \(\theta\).
\end{itemize}

We assume the volatility \(g_p(t,w_t)\) to be a constant matrix for each of the time regions $R_i$. For the function \(f_h(t,h_t;w_t)\), we use a similar architecture where the parameters correspond to the partially stochastic weights \(w_t\). This construction allows us to find suitable parameters for our PSDE-BNN which approximate $\tilde{f}$ with high probability. 

\begin{table*}[ht] 
\centering 
\caption{Classification accuracy and expected calibration error (ECE) on MNIST and CIFAR-10 (across 3 seeds). We borrow several baseline results from Xu et al. \yrcite{xu2022infinitely} (indicated by $\dag$) and add the results for our PSDE-BNN models and an extra ResNet32 + Last-Layer Laplace approximation. Models are separated into point estimates, stochastic and partially stochastic ``discrete-time", and continuous-time models. We also provide the SDE-BNN model performance when trained for the same amount of time it takes our models to train.} 
\label{tab: results}
\vspace{0.1in}
\begin{tabular}{@{}lcccc@{}}
\toprule
& \multicolumn{2}{c}{\textbf{MNIST}} & \multicolumn{2}{c}{\textbf{CIFAR-10}} \\
\cmidrule(lr){2-3} \cmidrule(lr){4-5}
\textbf{Model} & Accuracy (\%) $\uparrow$ & ECE (\(\times 10^{-2}\)) $\downarrow$ & Accuracy (\%) $\uparrow$ & ECE (\(\times 10^{-2}\)) $\downarrow$ \\
\midrule
ResNet32$^\dag$ & \textbf{99.46} \(\pm\) \textbf{0.00} & 2.88 \(\pm 0.94\) & 87.35 \(\pm 0.00\) & \textbf{8.47} \(\pm\) \textbf{0.39} \\
ODENet$^\dag$ & 98.90 \(\pm 0.04\) & \textbf{1.11} \(\pm\) \textbf{0.10} & \textbf{88.30} \(\pm\) \textbf{0.29} & 8.71 \(\pm 0.21\) \\
\midrule
MFVI ResNet 32$^\dag$ & \textbf{99.40} $\pm$ 0.00 & 2.76 $\pm$ 1.28 & 86.97 $\pm$ 0.00 & 3.04 $\pm$ 0.94\\
MFVI$^\dag$ & — & — & 86.48 & \textbf{1.95} \\
ResNet32 + LL Laplace & — & — & \textbf{92.10} & 2.98 \\
Deep Ensemble$^\dag$ & — & — & 89.22 & 2.79 \\
HMC (“gold standard”)$^\dag$ & 98.31 & \textbf{1.79} & 90.70 & 5.94 \\
\midrule
MFVI ODENet$^\dag$ & 98.81 \(\pm 0.00\) & 2.63 \(\pm 0.31\) & 81.59 \(\pm 0.01\) & 3.62 \(\pm 0.40\) \\
MFVI HyperODENet$^\dag$ & 98.77 \(\pm 0.01\) & 2.82 \(\pm 1.34\) & 80.62 \(\pm 0.00\) & 4.29 \(\pm 1.10\) \\
SDE-BNN & 98.55 \(\pm 0.09\) & 0.63 \(\pm 0.10\) & 86.04 \(\pm 0.25\) & 9.13 \(\pm 0.28\) \\
PSDE-BNN - ODEFirst (\emph{ours}) & \textbf{99.30} \(\pm\)\textbf{0.06} & 0.62 \(\pm 0.08\) & \textbf{87.84} \(\pm\) \textbf{0.08} & 4.73 \(\pm 0.07\) \\
PSDE-BNN - SDEFirst (\emph{ours}) & 99.10 \(\pm 0.07\) & \textbf{0.56} $\pm$ \textbf{0.10} & 85.34 \(\pm 0.21\) & \textbf{3.56} \(\pm\) \textbf{0.15}\\
PSDE-BNN - \texttt{fix\_w}$_2$ (\emph{ours}) & 99.30 \(\pm 0.07\) & 0.60 $\pm$ 0.10 & 87.49 \(\pm 0.24\) & 5.27 \(\pm\) 0.17 \\
PSDE-BNN - Hor. Cut (\emph{ours}) & 99.27 $\pm$ 0.03 & 0.57 $\pm$ 0.06 & 87.78 $\pm$ 0.37 & 4.29 $\pm$ 0.37 \\
\midrule
SDE-BNN (\emph{same training time}) & 94.23 \(\pm\) 0.02 & \textbf{0.14} \(\pm\) \textbf{0.02} & 69.94 \(\pm\) 0.21 & \textbf{1.52} \(\pm\) \textbf{0.23} \\
\bottomrule
\end{tabular}
\end{table*}

\section{Experiments}

In this section, we detail our Partially Stochastic Infinitely Deep BNNs' performance for image classification, uncertainty quantification, and learning efficiency. We compare these results with fully stochastic, partially stochastic and deterministic baselines, and we perform ablation studies to understand the impact of key hyperparameters.

\subsection{Experimental Setup}
In the design of our model, we implement a similar approach to Xu et al. \yrcite{xu2022infinitely} by parametrising the drift of the variational posterior $f_q$ through a multilayer perceptron. On the other hand, we choose a convolutional architecture for $f_h$, although it can be widely varied across the given tasks.

We conduct our experiments under three distinct configurations. The first configuration, termed `SDEfirst', is defined by setting the initial time cut, $t_1$, to 0. In this setup, the weights process is initially modelled with an SDE and then transitions to an ODE regime. Within this framework, we explore two sub-settings: one where the initial values of the ODE are the weights obtained from the SDE, and another, labelled `\texttt{fix\_w}$_2$', where the initial values of the ODE are trainable parameters. The second configuration, `ODEFirst', sets the second time cut, $t_2$, to 1, i.e. the endpoint of the considered time frame. Here, the weight evolution initially follows an ODE, with stochasticity introduced in the subsequent phase. The last configuration, the Horizontal Cut method (`Hor. Cut'), splits the weight vector into two subsets, $S$ and $D$ -  here $t_1=0$ and $t_2=1$. These segments evolve according to an SDE, discretised using the Euler-Maruyama scheme, and an ODE computed using the Euler scheme. Two critical criteria guide the selection of  $t_1$  and  $t_2$. The first criterion is the data’s uncertainty and complexity. For data with high uncertainty or limited prior knowledge, extending the stochastic phase ( $t_2$  in `SDEFirst') may be advantageous. Conversely, when more prior information is available, prolonging the deterministic phase ( $t_1$  in `ODEFirst') could benefit the model more. The second criterion is computational efficiency. Models with an extended stochastic phase may require greater computational resources, which is an important consideration in scenarios with limited computational budgets. Table \ref{tab: ablation} provides further insights into the impact of  $t_1$  and  $t_2$  on the performance and computational time of our models.

We employ a comprehensive grid search for hyperparameter optimisation, and the final values are detailed in Appendix \ref{app: experiments}. Across all experiments, the diffusion function $g_p$ is set to a constant value. Lastly, experiments were conducted on a single Nvidia RTX 3090 GPU.

\subsection{Image Classification}

Similar to Xu et al. \yrcite{xu2022infinitely}, the hidden state's instantaneous modifications ($f_h$) are parameterised with a convolutional neural network, which includes a strided convolution for downsampling and a transposed convolution layer for upsampling. The parameters $w_t$ are set to be the weights and biases of these layers. The MLP parametrising $f_q$ includes bottleneck layers (e.g. 2-128-2) in order to minimise the variational parameters and ensure linear scalability with the dimension of $w_t$. For the MNIST \cite{lecun2010mnist} dataset, a single SDE-BNN block is utilised, whereas, for CIFAR-10 \cite{Krizhevsky09learningmultiple}, we employ a multi-scale approach with several SDE-BNN blocks with the invertible downsampling from Dinh et al. \yrcite{dinh2016density} in between.

The classification outcomes, as detailed in Table \ref{tab: results}, demonstrate the superior performance of our model against baseline models across the datasets. Notably, while ODEnet architectures \citep{hu2020revealing} achieve classification metrics comparable to standard residual networks, they exhibit inferior calibration. In contrast, our PSDE-BNN model demonstrates superior accuracy relative to other stochastic frameworks, such as mean field variational inference (MFVI) ODENets, which utilize stochastic variational inference (SVI) over depth-invariant weights. Moreover, even in the domain of image classification, where discrete networks like ResNet32 with last-layer Laplace approximation perform robustly, our continuous-depth approach remains competitive. It's crucial to recognise the differences between these architectures; discrete networks may inherently align more closely with image classification tasks. However, our PSDE-BNN models achieve comparable performance without mechanisms like last-layer Laplace approximation, which are inapplicable in the infinite-depth setting. Moreover, the analysis reveals a consistent level of performance across the three distinct PSDE-BNN configurations, underlining the robustness of our approach regardless of the specific setting.


\subsection{Uncertainty Quantification}

\begin{figure*}[h!]  
    \centering
    \includegraphics[width=0.245\textwidth]{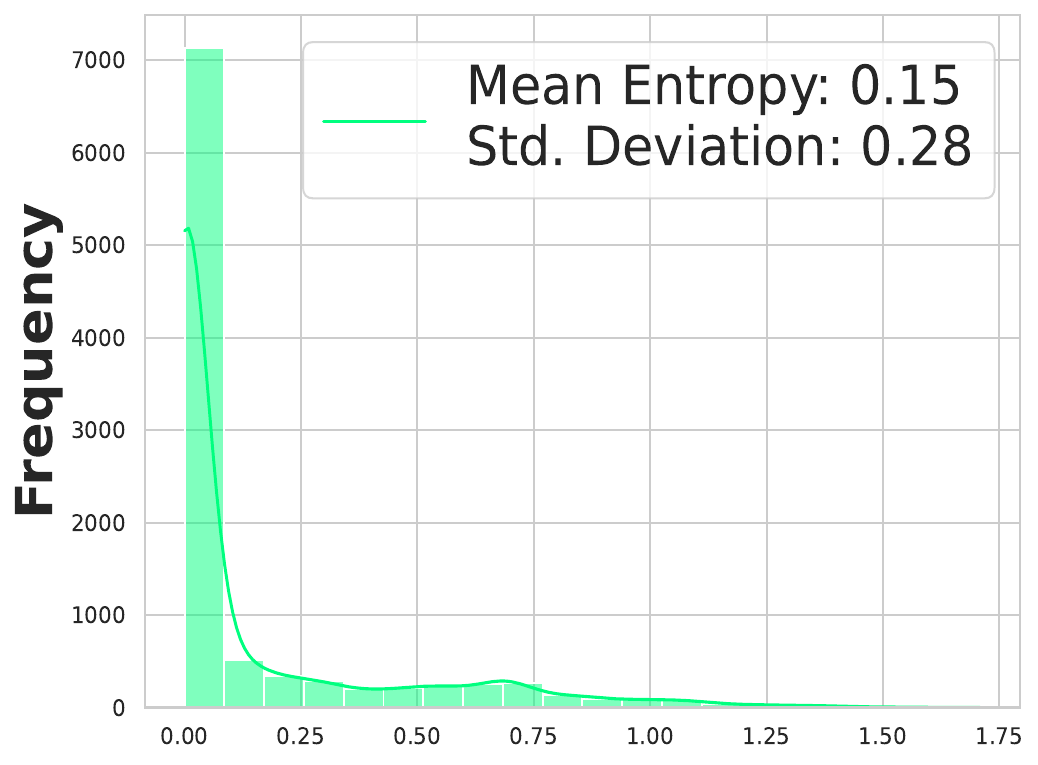}
    \includegraphics[width=0.245\textwidth]{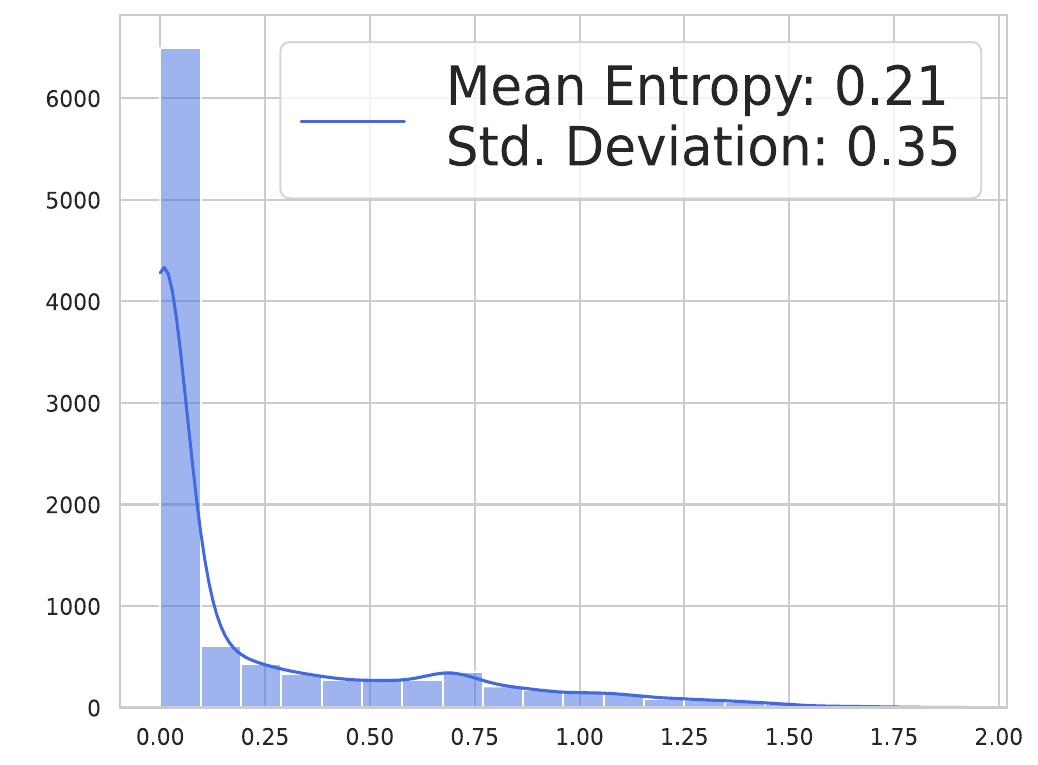}
    \includegraphics[width=0.245\textwidth]{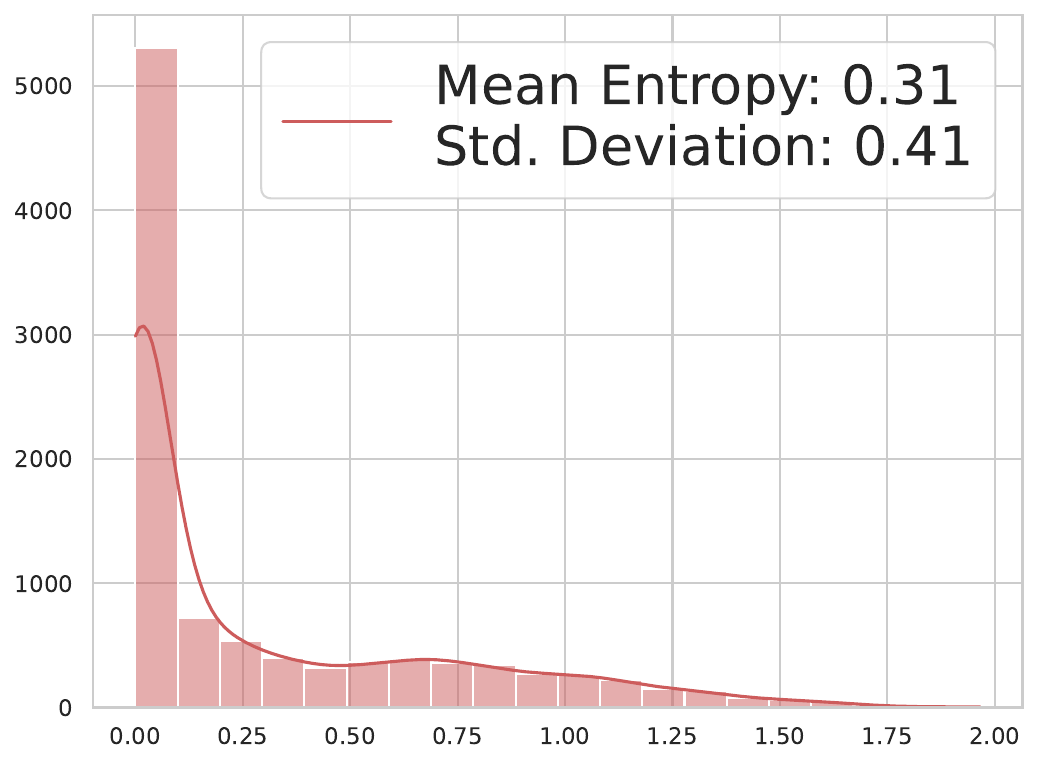}
    \includegraphics[width=0.245\textwidth]{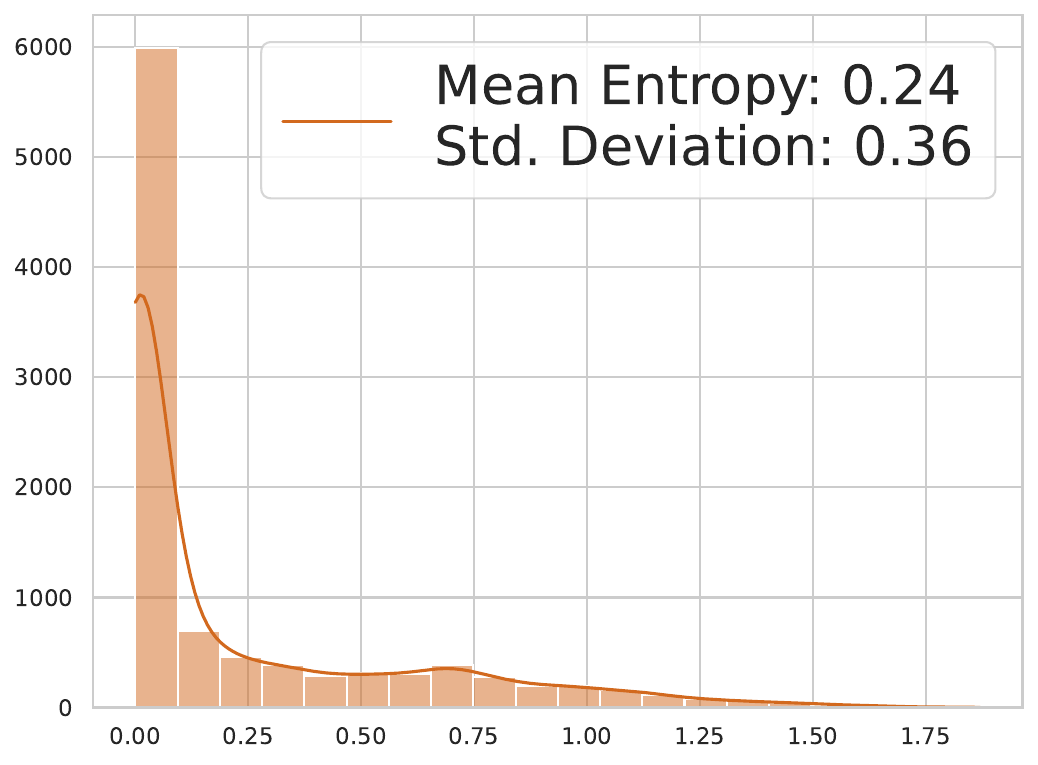}
    
    \includegraphics[width=0.245\textwidth]{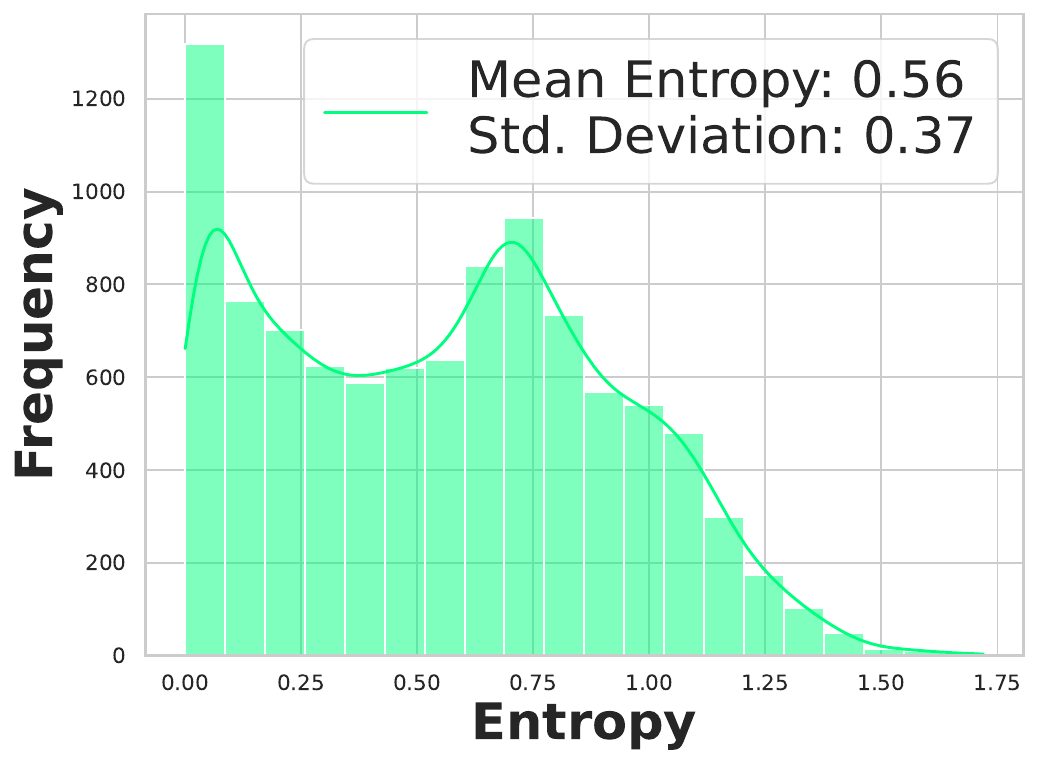}
    \includegraphics[width=0.245\textwidth]{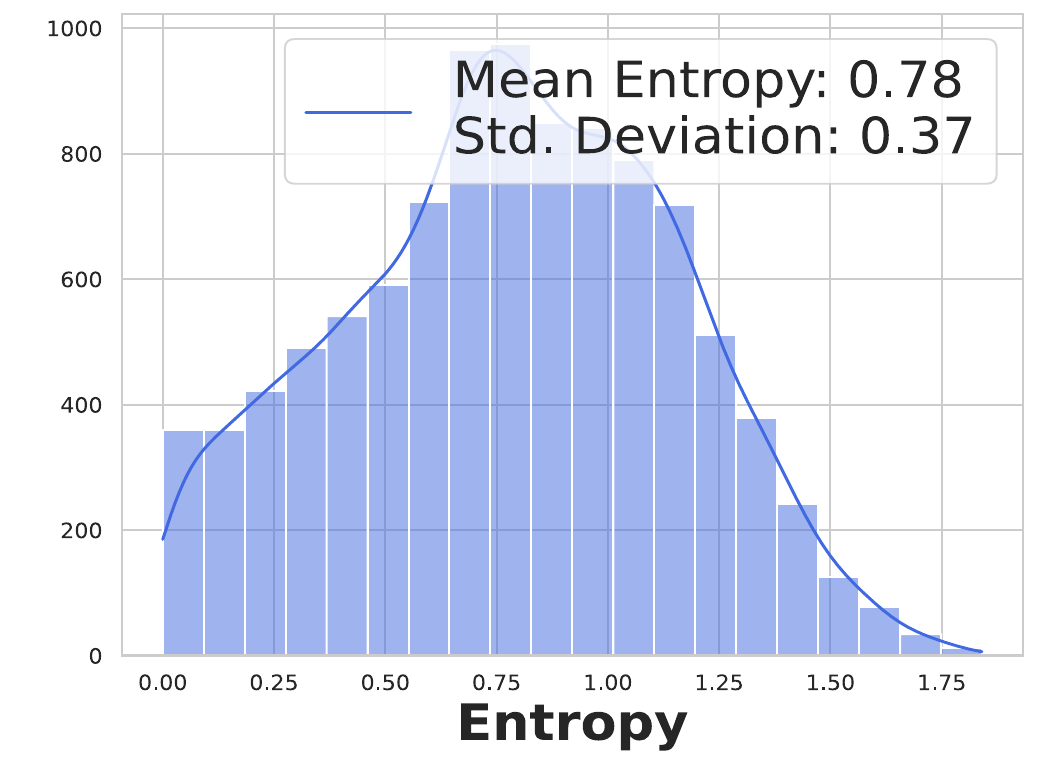}
    \includegraphics[width=0.245\textwidth]{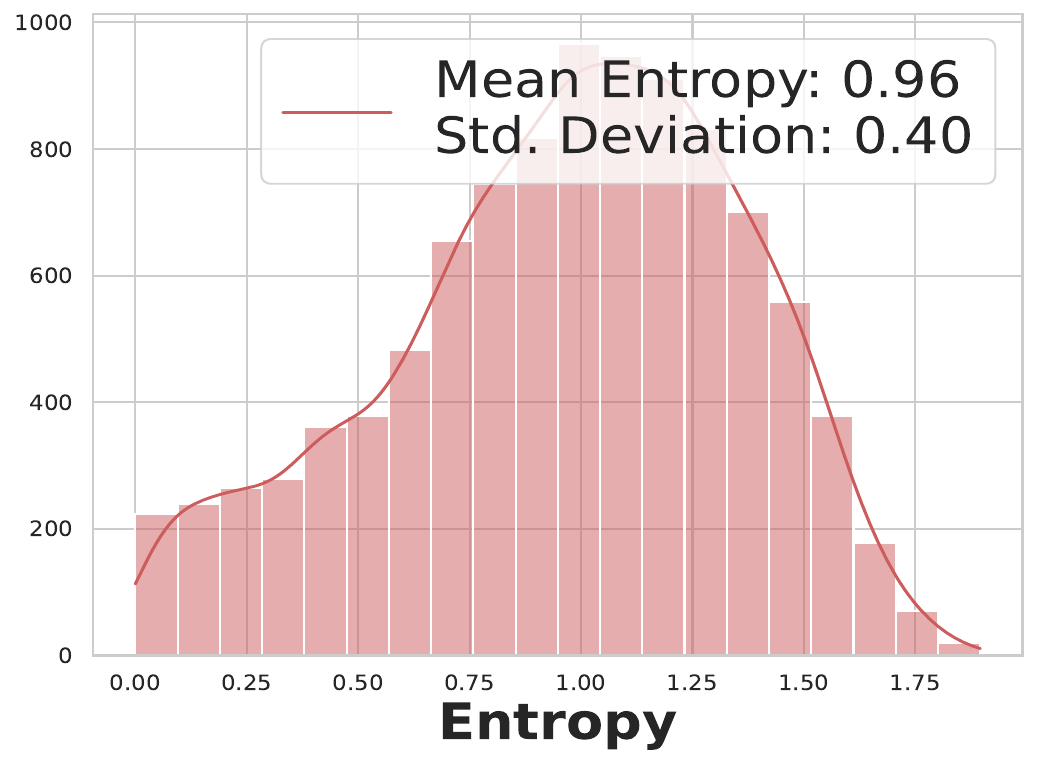}
    \includegraphics[width=0.248\textwidth]{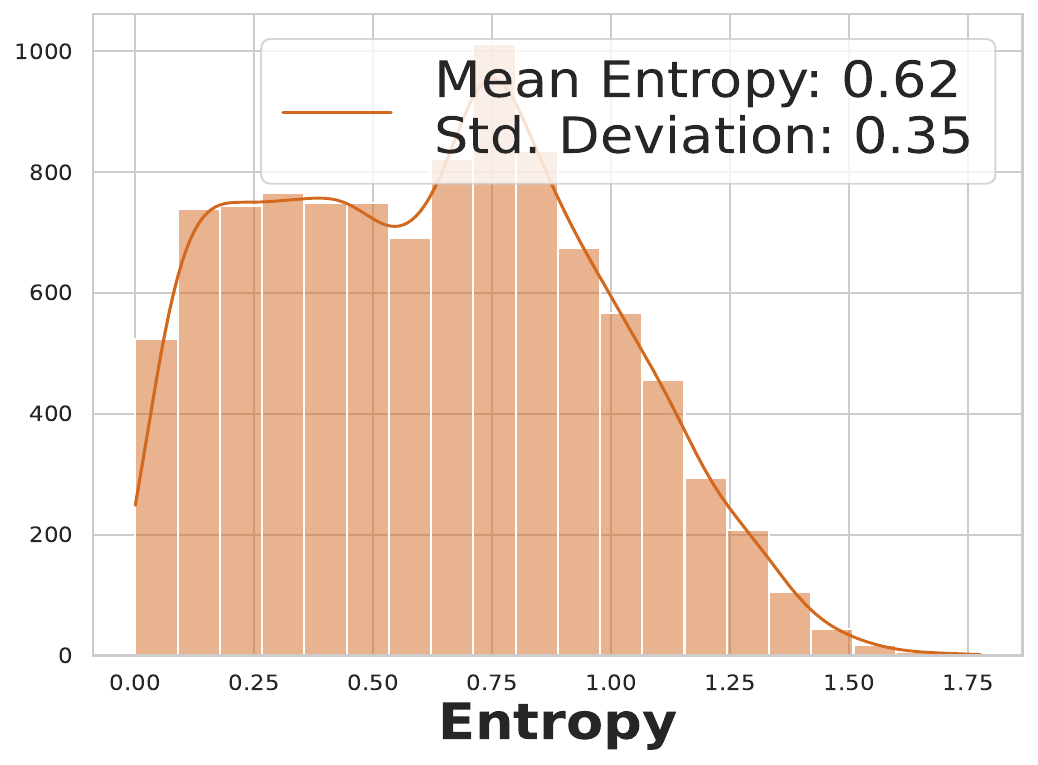}
    \vspace{-0.22in}
    \caption{Histograms depicting the distribution of predictive uncertainty for SDE-BNN (green) and PSDE-BNN ODEFirst (blue), SDEFirst (red), and with horizontal cuts (orange) across CIFAR-10 test predictions (top row) and OOD samples predictions (bottom row). Entropy was computed for each prediction, with higher values indicating greater uncertainty. Note that the legend includes summary statistics of the histograms, i.e. mean and standard deviation.}
    \label{fig: entropy-analysis}
\end{figure*}

\begin{table*}[h]
\centering
\begin{tabular}{lcccc}
\toprule
\textbf{Model}   & \textbf{Corruption Lvl 2} & \textbf{Corruption Lvl 3} & \textbf{Corruption Lvl 4} & \textbf{Corruption Lvl 5} \\ \midrule
ODEFirst         & \(0.322 \pm 0.409\)       & \(0.381 \pm 0.431\)       & \(0.431 \pm 0.450\)       & \(0.503 \pm 0.465\)       \\ 
SDEFirst         & \( \mathbf{0.443 \pm 0.468} \)      & \( \mathbf{0.498 \pm 0.480}\)       & \( \mathbf{0.543 \pm 0.489}\)      & \( \mathbf{0.628 \pm 0.512} \)      \\ 
Hor. Cut         & \(0.394 \pm 0.436\)       & \(0.430 \pm 0.447\)       & \(0.488 \pm 0.452\)       & \(0.541 \pm 0.458\)       \\ 
SDEBNN           & \(0.256 \pm 0.353\)       & \(0.293 \pm 0.370\)       & \(0.333 \pm 0.388\)       & \(0.375 \pm 0.405\)       \\ 
\bottomrule
\end{tabular}
\caption{Quantitative evaluation of model performance under varying corruption levels in the CIFAR10-C dataset. All models exhibit an increase in predictive entropy as the level of corruption intensifies, demonstrating their sensitivity to image degradation. Notably, models such as SDEBNN display a more pronounced increase in entropy across corruption levels, indicating a robust capability to recognize and respond to deteriorating input quality.}
\label{tab:corruption_levels}
\end{table*}

In the discrete neural networks domain, partial stochasticity is initially conceptualised as a simplified approximation of fully stochastic neural networks aimed at providing just enough uncertainty quantification. However, as we have theoretically established in earlier sections, partial stochasticity provides an expressive framework to approximate any posterior distribution. In this section, we further prove this empirically by looking at the expected calibration error (ECE) of the models at test time, and by performing predictive entropy analysis of out-of-distribution (OOD) and in-distribution (ID) samples.

Table \ref{tab: results} demonstrates that among the uncertainty-aware models, our PSDE-BNN (SDEFirst) achieves superior Expected Calibration Error (ECE) performance compared to all baseline models. Although ODEFirst, SDEFirst with ‘\texttt{fix\_w$_2$}’, and PSDE-BNN (Hor. Cut) consistently exhibit competitive error calibration and accuracy, the SDE-BNN encounters difficulties in maintaining low ECE when pursuing higher accuracy. This trade-off arises during training, where ECE is compromised for enhanced accuracy, likely due to overfitting that progressively narrows uncertainty estimates. This phenomenon is particularly pronounced in the SDE-BNN. For CIFAR-10, at relatively low accuracies (70\% - 80\%), SDE-BNN can achieve good ECE performance, but this quickly degrades over epochs, and at a much faster rate than PSDE-BNNs. This is shown in the last row of Table \ref{tab: results}, where SDE-BNN, within the same time limit, achieves excellent ECE at the cost of reduced accuracy, which would improve with additional epochs.

Encouraged by the calibration results, we proceed to evaluate model performance in distinguishing in-distribution (ID) from out-of-distribution (OOD) samples. A well-known issue with deterministic networks is their tendency to be overly confident in their predictions, regardless of the input. Ideally, models with effective uncertainty quantification should mitigate this problem for OOD samples. Our evaluation, based on ROC curves and AUC values (see Appendix \ref{app: roc-curves}), demonstrates that our partially stochastic models (AUC: 0.88 for models with vertical cuts, AUC: 0.86 for the model with horizontal cut) outperform the SDE-BNN model (AUC: 0.84). Figure \ref{fig: entropy-analysis} further analyzes predictive entropy distributions, highlighting our models’ superior performance in representing OOD uncertainty.

Moreover, we utilized CIFAR10-C \cite{hendrycks2019robustness}, a dataset that introduces various corruption levels to systematically assess our models’ uncertainty estimation. Our investigations involved adding 17 different types of corruptions (e.g., Gaussian noise, blur, snow, compression, pixelation) at five different intensity levels to challenge the models’ robustness and generalization. The results are presented in Table \ref{tab:corruption_levels}.

\subsection{Efficiency and Learning Rate}
The approach from Xu et al. \yrcite{xu2022infinitely}, has limited use in practice due to excessively long training and inference times. Indeed, one of the main incentives behind this paper is to provide an efficient alternative to SDE-BNN that is practical and can be efficiently used for downstream tasks. To determine the efficiency of our models, we measure the average time taken to compute an epoch and the inference time required to make 10000 CIFAR-10 predictions. In this section, we focus on models with vertical cuts due to the current lack of an optimised numerical solver for the horizontal cut model. Table \ref{tab:model_times} shows a significant gap between both PSDE-BNN models and SDE-BNN in both tasks. Our faster model achieves \textbf{27.5\% quicker epoch} completion on average and is \textbf{25.9\% faster at inference} than the baseline.


\begin{table}[htbp]
\centering
\caption{Models' average times in seconds.}
\label{tab:model_times}
\begin{tabular}{
  l
  S[table-format=3.1] 
  S[table-format=3.1] 
}
\toprule
{\textbf{Model}} & {\textbf{Epoch Time}} & {\textbf{Inference Time}} \\
\midrule
SDE-BNN & 371.9 & 43.7 \\ 
ODEFirst (10\%) & \textbf{289.4} & \hspace{0.07in}\textbf{31.7} \\ 
SDEFirst (10\%) & 295.1 & \hspace{0.07in}\textbf{31.7} \\ 
Hor. Cut (50\%) & 316.7 & \hspace{0.07in}{35.9} \\ %
\bottomrule
\end{tabular}
\vspace{-0.2in}
\end{table}

Next, we explore the learning rate of our models, i.e. how many epochs the network takes to get to acceptable downstream task performance. PSDE-BNN takes considerably less time to achieve the reported classification accuracies. While our models trained for 100 epochs, SDE-BNN (and as reported by Xu et al. \yrcite{xu2022infinitely}) required 300 epochs. The evolution after the 80\% test accuracy for SDE-BNN was especially long. This indicates that PSDE-BNNs were both faster at each epoch, and only needed a third of the epochs to achieve similar (although better) downstream performance. Therefore, the overall training time was \textbf{74.1\% faster}.

\subsection{Ablation Studies}
Here, we explore the sensitivity of the PSDE-BNN model with a vertical separation of the weights to key hyperparameters. We determine the impact of the stochasticity ratio, which modulates the duration of stochastic dynamics in our framework, and the KL coefficient ($\kappa$ in  \eqref{eq:elbo}) on model performance.

The ablation studies (Table \ref{tab: ablation} in Appendix \ref{app: train-plots}), show no correlation between the stochasticity ratio and downstream performance or uncertainty metrics (ECEs and entropy). This is consistent with the literature in partially stochastic networks, where several approaches achieve competitive uncertainty representations with only the last discrete layer of the network being stochastic (e.g. using Laplace approximation or HMC) demonstrating that a small subset of stochastic weights suffices (\citealt{daxberger2022laplace, duranmartin2021efficient, immer2021improving, antorán2023samplingbased}). However, higher stochasticity ratios increase computation times. Given the lack of performance trade-off, the stochasticity ratio in our optimal model is set to 10\%. Moreover, we find that increasing the order of $\kappa$ to $10^{-2}$ leads to an enhanced learning rate but at the cost of numerical instability during training. In contrast, decreasing the order of $\kappa$ to $10^{-4}$ makes the training of the models converge slower. Given this trade-off, we conclude that the default value of $\kappa = 10^{-3}$ is the most versatile (see Appendix \ref{app: train-plots}).

\vspace{-0.1in}
\section{Conclusion and Future Work}

We have introduced Partially Stochastic Bayesian Neural Networks (PSDE-BNNs) in the infinite-depth limit, providing an exact mathematical formulation and demonstrating how to partition the weights of continuous-time networks into stochastic and deterministic subsets. Building on these foundations, we established several theoretical results that lead to the conclusion that our novel family of networks are Universal Conditional Distribution Approximators. Our empirical experiments show that PSDE-BNNs outperform fully stochastic networks and other Bayesian inference approximations in continuous-time models, such as Mean Field Variational Inference. Notably, this enhanced performance is achieved with a significant efficiency gain, without any trade-off in performance. Additionally, we conducted an investigation of the effects of the key hyperparameters in our models. In summary, our newly introduced family of Bayesian Neural Networks in the infinite-depth limit offers a superior, more efficient, and practical alternative to existing models.

In future work, we aim to investigate models that incorporate both horizontal and vertical cuts to explore potential performance improvements. Additionally, we will explore variance reduction techniques such as Sticking The Landing \citep{roeder2017stl} to enhance the performance and stability of our models. Another promising direction is to evaluate the generalization capabilities of our approaches on a diverse range of datasets, including Omniglot for few-shot learning and ImageNet for high-resolution performance. Given that scalability is a common challenge with Bayesian models, implementing PSDE-BNN on learned features may further improve model efficiency. Finally, we plan to systematically investigate the relationship between test set performance and the quality of the variational approximation. This includes exploring different priors and model complexities to better understand how the KL divergence impacts generalization across various settings.

\section*{Impact Statement}

The impact of our work manifests primarily in the domain of model uncertainty quantification. This work does not specifically focus on any particular ethical challenge that we should worry about nowadays such as LLMs or image generation models. However, advanced predictive models can influence a wide range of applications, from healthcare to autonomous systems. Therefore, we encourage ongoing reflection on how advancements in machine learning may shape and be shaped by societal and ethical considerations.

\section*{Acknowledgements}

We thank the CDT of Mathematics in Random Systems for backing the entirety of our research team. SCO and YS also extend gratitude to the Oxford-Man Institute of Quantitative Finance for their support. The manuscript has benefited from the insightful feedback of H. Waldon, F. Drissi, and J. He, for which all authors are appreciative. SCO would also like to thank R. Bergna for the stimulating discussions.

\bibliography{example_paper}

\begin{thebibliography}{49}
\providecommand{\natexlab}[1]{#1}
\providecommand{\url}[1]{\texttt{#1}}
\expandafter\ifx\csname urlstyle\endcsname\relax
  \providecommand{\doi}[1]{doi: #1}\else
  \providecommand{\doi}{doi: \begingroup \urlstyle{rm}\Url}\fi

\bibitem[Antorán et~al.(2023)Antorán, Padhy, Barbano, Nalisnick, Janz, and Hernández-Lobato]{antorán2023samplingbased}
Antorán, J., Padhy, S., Barbano, R., Nalisnick, E., Janz, D., and Hernández-Lobato, J.~M.
\newblock Sampling-based inference for large linear models, with application to linearised laplace.
\newblock \emph{arXiv preprint arXiv:2210.04994}, 2023.

\bibitem[Archambeau et~al.(2007)Archambeau, Opper, Shen, Cornford, and Shawe-Taylor]{archambeau2007variational}
Archambeau, C., Opper, M., Shen, Y., Cornford, D., and Shawe-Taylor, J.
\newblock Variational inference for diffusion processes.
\newblock \emph{Advances in neural information processing systems}, 20, 2007.

\bibitem[Arroyo et~al.(2024)Arroyo, Cartea, Moreno-Pino, and Zohren]{arroyo2024deep}
Arroyo, A., Cartea, A., Moreno-Pino, F., and Zohren, S.
\newblock Deep attentive survival analysis in limit order books: Estimating fill probabilities with convolutional-transformers.
\newblock \emph{Quantitative Finance}, pp.\  1--23, 2024.

\bibitem[Bergna et~al.(2023)Bergna, Opolka, Liò, and Hernandez-Lobato]{bergna2023graph}
Bergna, R., Opolka, F., Liò, P., and Hernandez-Lobato, J.~M.
\newblock Graph neural stochastic differential equations.
\newblock \emph{arXiv preprint arXiv:2308.12316}, 2023.

\bibitem[Blundell et~al.(2015)Blundell, Cornebise, Kavukcuoglu, and Wierstra]{blundell2015weight}
Blundell, C., Cornebise, J., Kavukcuoglu, K., and Wierstra, D.
\newblock Weight uncertainty in neural networks.
\newblock In \emph{Proceedings of the 32nd International Conference on Machine Learning}, volume~37, pp.\  1613--1622, 2015.

\bibitem[Calvo-Ordoñez et~al.(2023)Calvo-Ordoñez, Cheng, Huang, Zhang, Yang, Schonlieb, and Aviles-Rivero]{calvoordonez2023missing}
Calvo-Ordoñez, S., Cheng, C.-W., Huang, J., Zhang, L., Yang, G., Schonlieb, C.-B., and Aviles-Rivero, A.~I.
\newblock The missing u for efficient diffusion models.
\newblock \emph{arXiv preprint arXiv:2310.20092}, 2023.

\bibitem[Cattiaux \& L{\'e}onard(1994)Cattiaux and L{\'e}onard]{cattiaux1994minimization}
Cattiaux, P. and L{\'e}onard, C.
\newblock Minimization of the kullback information of diffusion processes.
\newblock In \emph{Annales de l'IHP Probabilit{\'e}s et statistiques}, volume~30, pp.\  83--132, 1994.

\bibitem[Chen et~al.(2018)Chen, Rubanova, Bettencourt, and Duvenaud]{Neural_ODEs}
Chen, R. T.~Q., Rubanova, Y., Bettencourt, J., and Duvenaud, D.~K.
\newblock Neural ordinary differential equations.
\newblock In \emph{Advances in Neural Information Processing Systems}, volume~31, 2018.

\bibitem[Daxberger et~al.(2021{\natexlab{a}})Daxberger, Kristiadi, Immer, Eschenhagen, Bauer, and Hennig]{daxberger2022laplace}
Daxberger, E., Kristiadi, A., Immer, A., Eschenhagen, R., Bauer, M., and Hennig, P.
\newblock Laplace redux-effortless bayesian deep learning.
\newblock In \emph{Advances in Neural Information Processing Systems}, volume~34, pp.\  20089--20103, 2021{\natexlab{a}}.

\bibitem[Daxberger et~al.(2021{\natexlab{b}})Daxberger, Nalisnick, Allingham, Antor{\'a}n, and Hern{\'a}ndez-Lobato]{daxberger2022bayesian}
Daxberger, E., Nalisnick, E., Allingham, J.~U., Antor{\'a}n, J., and Hern{\'a}ndez-Lobato, J.~M.
\newblock Bayesian deep learning via subnetwork inference.
\newblock In \emph{Proceedings of the 38th International Conference on Machine Learning}, volume 139, pp.\  2510--2521, 2021{\natexlab{b}}.

\bibitem[Dinh et~al.(2016)Dinh, Sohl-Dickstein, and Bengio]{dinh2016density}
Dinh, L., Sohl-Dickstein, J., and Bengio, S.
\newblock Density estimation using real nvp.
\newblock \emph{arXiv preprint arXiv:1605.08803}, 2016.

\bibitem[Dupont et~al.(2019)Dupont, Doucet, and Teh]{dupont2019augmented}
Dupont, E., Doucet, A., and Teh, Y.~W.
\newblock Augmented neural odes.
\newblock In \emph{Advances in Neural Information Processing Systems}, volume~32, 2019.

\bibitem[Duran-Martin et~al.(2022)Duran-Martin, Kara, and Murphy]{duranmartin2021efficient}
Duran-Martin, G., Kara, A., and Murphy, K.
\newblock Efficient online bayesian inference for neural bandits.
\newblock In \emph{Proceedings of the 25th International Conference on Artificial Intelligence and Statistics}, volume 151, pp.\  6002--6021, 2022.

\bibitem[Farquhar et~al.(2020)Farquhar, Smith, and Gal]{gal_liberty_or_depth}
Farquhar, S., Smith, L., and Gal, Y.
\newblock Liberty or depth: Deep bayesian neural nets do not need complex weight posterior approximations.
\newblock In \emph{Advances in Neural Information Processing Systems}, volume~33, pp.\  4346--4357, 2020.

\bibitem[Ghosh et~al.(2022)Ghosh, Birrell, and De~Angelis]{ghosh2022differentiable}
Ghosh, S., Birrell, P.~J., and De~Angelis, D.
\newblock Differentiable bayesian inference of sde parameters using a pathwise series expansion of brownian motion.
\newblock \emph{Proceedings of The 25th International Conference on Artificial Intelligence and Statistics}, pp.\  10982--10998, 2022.

\bibitem[Haber \& Ruthotto(2017)Haber and Ruthotto]{Haber_2017}
Haber, E. and Ruthotto, L.
\newblock Stable architectures for deep neural networks.
\newblock \emph{Inverse problems}, 34\penalty0 (1):\penalty0 014004, 2017.

\bibitem[Harrison et~al.(2024)Harrison, Willes, and Snoek]{harrison2024variational}
Harrison, J., Willes, J., and Snoek, J.
\newblock Variational bayesian last layers.
\newblock \emph{arXiv preprint arXiv:2404.11599}, 2024.

\bibitem[He et~al.(2023)He, Flamich, Guo, and Hern{\'a}ndez-Lobato]{he2023recombiner}
He, J., Flamich, G., Guo, Z., and Hern{\'a}ndez-Lobato, J.~M.
\newblock Recombiner: Robust and enhanced compression with bayesian implicit neural representations.
\newblock \emph{arXiv preprint arXiv:2309.17182}, 2023.

\bibitem[He et~al.(2016)He, Zhang, Ren, and Sun]{he2016deep}
He, K., Zhang, X., Ren, S., and Sun, J.
\newblock Deep residual learning for image recognition.
\newblock In \emph{Proceedings of the IEEE conference on computer vision and pattern recognition}, pp.\  770--778, 2016.

\bibitem[Hendrycks \& Dietterich(2019)Hendrycks and Dietterich]{hendrycks2019robustness}
Hendrycks, D. and Dietterich, T.
\newblock Benchmarking neural network robustness to common corruptions and perturbations.
\newblock \emph{Proceedings of the International Conference on Learning Representations}, 2019.

\bibitem[Hoffman et~al.(2013)Hoffman, Blei, Wang, and Paisley]{hoffman13-svi}
Hoffman, M.~D., Blei, D.~M., Wang, C., and Paisley, J.
\newblock Stochastic variational inference.
\newblock \emph{Journal of Machine Learning Research}, 14\penalty0 (40):\penalty0 1303--1347, 2013.

\bibitem[Hu et~al.(2020)Hu, Yang, Zhu, and Hong]{hu2020revealing}
Hu, P., Yang, W., Zhu, Y., and Hong, L.
\newblock Revealing hidden dynamics from time-series data by odenet.
\newblock \emph{arXiv preprint arXiv:2005.04849}, 2020.

\bibitem[Immer et~al.(2021)Immer, Korzepa, and Bauer]{immer2021improving}
Immer, A., Korzepa, M., and Bauer, M.
\newblock Improving predictions of bayesian neural nets via local linearization.
\newblock In \emph{Proceedings of The 24th International Conference on Artificial Intelligence and Statistics}, volume 130, pp.\  703--711, 2021.

\bibitem[Kallenberg(2010)]{kallenberg_foundations_2010}
Kallenberg, O.
\newblock \emph{Foundations of {Modern} {Probability}}.
\newblock Probability and its {Applications}. Springer, New York, NY Berlin Heidelberg, 2. ed edition, 2010.
\newblock ISBN 9781441929495.

\bibitem[Kristiadi et~al.(2020)Kristiadi, Hein, and Hennig]{kristiadi2020bayesian}
Kristiadi, A., Hein, M., and Hennig, P.
\newblock Being bayesian, even just a bit, fixes overconfidence in relu networks.
\newblock In \emph{Proceedings of the 37th International Conference on Machine Learning}, volume 119, pp.\  5436--5446, 2020.

\bibitem[Krizhevsky(2009)]{Krizhevsky09learningmultiple}
Krizhevsky, A.
\newblock Learning multiple layers of features from tiny images, 2009.

\bibitem[Lampinen \& Vehtari(2001)Lampinen and Vehtari]{lampinen2001bayesian}
Lampinen, J. and Vehtari, A.
\newblock Bayesian approach for neural networks—review and case studies.
\newblock \emph{Neural Networks}, 14\penalty0 (3):\penalty0 257--274, 2001.

\bibitem[LeCun et~al.(2010)LeCun, Cortes, and Burges]{lecun2010mnist}
LeCun, Y., Cortes, C., and Burges, C.
\newblock Mnist handwritten digit database.
\newblock \emph{ATT Labs [Online]. Available: http://yann.lecun.com/exdb/mnist}, 2, 2010.

\bibitem[Leshno et~al.(1993)Leshno, Lin, Pinkus, and Schocken]{leshno93}
Leshno, M., Lin, V.~Y., Pinkus, A., and Schocken, S.
\newblock Multilayer feedforward networks with a nonpolynomial activation function can approximate any function.
\newblock \emph{Neural Networks}, 6\penalty0 (6):\penalty0 861--867, 1993.

\bibitem[Li et~al.(2024)Li, Miao, Qiu, and Zhang]{li2024training}
Li, J., Miao, Z., Qiu, Q., and Zhang, R.
\newblock Training bayesian neural networks with sparse subspace variational inference.
\newblock \emph{arXiv preprint arXiv:2402.11025}, 2024.

\bibitem[Li et~al.(2020)Li, Wong, Chen, and Duvenaud]{scalable_gradients_for_SDE}
Li, X., Wong, T.-K.~L., Chen, R. T.~Q., and Duvenaud, D.
\newblock Scalable gradients for stochastic differential equations.
\newblock In \emph{Proceedings of the 23rd International Conference on Artificial Intelligence and Statistics}, volume 108, pp.\  3870--3882, 2020.

\bibitem[Liu et~al.(2019)Liu, Xiao, Si, Cao, Kumar, and Hsieh]{liu2019neural}
Liu, X., Xiao, T., Si, S., Cao, Q., Kumar, S., and Hsieh, C.-J.
\newblock Neural sde: Stabilizing neural ode networks with stochastic noise.
\newblock \emph{arXiv preprint arXiv:1906.02355}, 2019.

\bibitem[Mackay(1992)]{article-mackay}
Mackay, D. J.~C.
\newblock \emph{Bayesian methods for adaptive models}.
\newblock PhD thesis, California Institute of Technology, 1992.

\bibitem[Moreno-Pino \& Zohren(2022)Moreno-Pino and Zohren]{moreno2022deepvol}
Moreno-Pino, F. and Zohren, S.
\newblock Deepvol: Volatility forecasting from high-frequency data with dilated causal convolutions.
\newblock \emph{arXiv preprint arXiv:2210.04797}, 2022.

\bibitem[Neal(1995)]{Neal1995BayesianLF}
Neal, R.~M.
\newblock \emph{Bayesian Learning For Neural Networks}.
\newblock PhD thesis, University of Toronto, 1995.

\bibitem[Oksendal(2013)]{oksendal2013stochastic}
Oksendal, B.
\newblock \emph{Stochastic differential equations: an introduction with applications}.
\newblock Springer Science \& Business Media, 2013.

\bibitem[Richter et~al.(2024)Richter, Berner, and Liu]{richter2024improved}
Richter, L., Berner, J., and Liu, G.-H.
\newblock Improved sampling via learned diffusions.
\newblock \emph{Proceedings of the International Conference on Learning Representations}, 2024.

\bibitem[Roeder et~al.(2017)Roeder, Wu, and Duvenaud]{roeder2017stl}
Roeder, G., Wu, Y., and Duvenaud, D.~K.
\newblock Sticking the landing: Simple, lower-variance gradient estimators for variational inference.
\newblock In \emph{Advances in Neural Information Processing Systems}, volume~30, 2017.

\bibitem[Sharma et~al.(2023)Sharma, Farquhar, Nalisnick, and Rainforth]{do_BNN_need_fully_stochastic}
Sharma, M., Farquhar, S., Nalisnick, E., and Rainforth, T.
\newblock Do bayesian neural networks need to be fully stochastic?
\newblock In \emph{Proceedings of The 26th International Conference on Artificial Intelligence and Statistics}, volume 206, pp.\  7694--7722, 2023.

\bibitem[Titterington(2004)]{titterington2004bayesian}
Titterington, D.~M.
\newblock Bayesian methods for neural networks and related models.
\newblock \emph{Statistical Science}, 19\penalty0 (1):\penalty0 128--139, February 2004.

\bibitem[Tzen \& Raginsky(2019)Tzen and Raginsky]{tzen2019neural}
Tzen, B. and Raginsky, M.
\newblock Neural stochastic differential equations: Deep latent gaussian models in the diffusion limit.
\newblock \emph{arXiv preprint arXiv:1905.09883}, 2019.

\bibitem[Ustyuzhaninov et~al.(2020)Ustyuzhaninov, Kazlauskaite, Kaiser, Bodin, Campbell, and Henrik~Ek]{pmlr-v124-ustyuzhaninov20a}
Ustyuzhaninov, I., Kazlauskaite, I., Kaiser, M., Bodin, E., Campbell, N., and Henrik~Ek, C.
\newblock Compositional uncertainty in deep gaussian processes.
\newblock In \emph{Proceedings of the 36th Conference on Uncertainty in Artificial Intelligence (UAI)}, volume 124, pp.\  480--489, 2020.

\bibitem[Van~Amersfoort et~al.(2020)Van~Amersfoort, Smith, Teh, and Gal]{vanamersfoort2020uncertainty}
Van~Amersfoort, J., Smith, L., Teh, Y.~W., and Gal, Y.
\newblock Uncertainty estimation using a single deep deterministic neural network.
\newblock In \emph{Proceedings of the 37th International Conference on Machine Learning}, volume 119, pp.\  9690--9700, 2020.

\bibitem[Vanderschueren et~al.(2023)Vanderschueren, Curth, Verbeke, and van~der Schaar]{vanderschueren2023accounting}
Vanderschueren, T., Curth, A., Verbeke, W., and van~der Schaar, M.
\newblock Accounting for informative sampling when learning to forecast treatment outcomes over time.
\newblock \emph{arXiv preprint arXiv:2306.04255}, 2023.

\bibitem[Vargas et~al.(2023)Vargas, Ovsianas, Fernandes, Girolami, Lawrence, and N{\"u}sken]{vargas2023bayesian}
Vargas, F., Ovsianas, A., Fernandes, D., Girolami, M., Lawrence, N.~D., and N{\"u}sken, N.
\newblock Bayesian learning via neural schr{\"o}dinger--f{\"o}llmer flows.
\newblock \emph{Statistics and Computing}, 33\penalty0 (1):\penalty0 3, 2023.

\bibitem[Vargas et~al.(2024)Vargas, Padhy, Blessing, and Nüsken]{vargas2024transport}
Vargas, F., Padhy, S., Blessing, D., and Nüsken, N.
\newblock Transport meets variational inference: Controlled monte carlo diffusions.
\newblock \emph{Proceedings of the International Conference on Learning Representations}, 2024.

\bibitem[Xu et~al.(2022)Xu, Chen, Li, and Duvenaud]{xu2022infinitely}
Xu, W., Chen, R.~T., Li, X., and Duvenaud, D.
\newblock Infinitely deep bayesian neural networks with stochastic differential equations.
\newblock In \emph{Proceedings of The 25th International Conference on Artificial Intelligence and Statistics}, volume 151, pp.\  721--738, 2022.

\bibitem[Yarotsky(2017)]{yarotsky17}
Yarotsky, D.
\newblock Error bounds for approximations with deep relu networks.
\newblock \emph{Neural Networks}, 94:\penalty0 103--114, 2017.
\newblock ISSN 0893-6080.

\bibitem[Zhang et~al.(2020)Zhang, Gao, Unterman, and Arodz]{zhang20h}
Zhang, H., Gao, X., Unterman, J., and Arodz, T.
\newblock Approximation capabilities of neural {ODE}s and invertible residual networks.
\newblock In \emph{Proceedings of the 37th International Conference on Machine Learning}, volume 119, pp.\  11086--11095, 2020.

\end{thebibliography}
\bibliographystyle{icml2024}

\newpage
\appendix
\onecolumn
\section{Proofs}\label{Appendix:proof}

\subsection{Proof of Theorem \ref{theo:non-approximate}}
    We know that the sample path of $w_t$ is almost surely continuous as the Brownian motion $B_t$ is almost surely continuous. For any $\omega\in\Omega$ such that $w_t$ is continuous, we can write $w_t$ as a continuous function of time $t$. Therefore, given the continuous sample path $w_t$, we can write $f_h(t,h_t;w_t)=f_h(t,h_t)$ where $f_h$ is continuous in $t$ and globally Lipschitz continuous in $h_t$. In this case, $h_t$ satisfies an ODE and we can follow the same argument as in Dupont et al. \yrcite{dupont2019augmented} and show that for any $\omega$ such that $w_t$ is continuous, $\mathcal{L}(\psi_1(\omega,x))$ cannot represent $z(x)$. Actually, following the same argument we can show that  $\mathcal{L}(\psi_1(\omega,x))$ cannot map $\{\|{x}\| \leq r_1,x\in\mathbb{R}^{d_x}\}$ to the region $\{x\leq-1/2,x\in\mathbb{R}\}$ and $\{r_2 \leq\|\mathbf{x}\| \leq r_3,x\in\mathbb{R}^{d_x}\}$ to the region $\{x\geq1/2,x\in\mathbb{R}\}$ simultaneously. If that is the case then $\psi_1(\{\|{x}\| \leq r_1,x\in\mathbb{R}^{d_x}\})$ and $\psi_1(\{r_2 \leq\|\mathbf{x}\| \leq r_3,x\in\mathbb{R}^{d_x}\})$ would be linearly separable which contradicts proposition 2 in Dupont et al. \yrcite{dupont2019augmented}. Therefore, if the sample path $w_t$ is continuous, $\sup_{x\in A}|\mathcal{L}(\psi_1(\omega,x))-z(x)|>1/2$ and we have:
$$
\mathbb{P}\left(\sup_{x\in A}|\mathcal{L}(\psi_1(\omega,x))-z(x)|>1/2\right)=\mathbb{P}(w_t\text{ is continuous})=1
$$   
\qed 

\vspace{-0.15in}
\subsection{Proof of Theorem \ref{thm:main-theorem}}
First we state the noise outsourcing lemma \cite{kallenberg_foundations_2010}, two theorems by Zhang et al. \yrcite{zhang20h} and the theorem by Sharma et al. \yrcite{do_BNN_need_fully_stochastic}:
\begin{lemma}
\label{lemma:noise-outsourcing}
   (Noise Outsourcing Lemma \cite{kallenberg_foundations_2010}). Let $X$ and $Y$ be random variables in Borel spaces $\mathcal{X}$ and $\mathcal{Y}$. For any given $m \geq 1$, there exists a random variable $\eta \sim \mathcal{N}\left(0, I_m\right)$ and a Borel-measurable function $\tilde{f}: \mathbb{R}^m \times \mathcal{X} \rightarrow \mathcal{Y}$ such that $\eta$ is independent of $X$ and
$$
(X, Y)=(X, \tilde{f}(\eta, X))
$$
almost surely. Thus, $\tilde{f}(\eta, x) \sim Y \mid X=x, \forall x \in \mathcal{X}$. 
\end{lemma}

\begin{lemma}
    \label{lemma:homeo-approx-net}
    (Theorem 2 by Zhang et al. \yrcite{zhang20h}). For any homeomorphism $h: \mathcal{X} \rightarrow \mathcal{X}, \mathcal{X} \subset \mathbb{R}^p$, there exists a $2 p$-ODE-Net $\phi_T: \mathbb{R}^{2 p} \rightarrow \mathbb{R}^{2 p}$ for $T=1$ such that $\phi_T\left(\left[x, 0^{(p)}\right]\right)=\left[h(x), 0^{(p)}\right]$ for any $x \in \mathcal{X}$.
\end{lemma}

\begin{lemma}
    \label{lemma:node-approx-net}
    (Theorem 7 by Zhang et. al \yrcite{zhang20h}). Consider a neural network $F: \mathbb{R}^p \rightarrow \mathbb{R}^r$. For $q=p+r$, there exists a linear layer-capped $q$-ODE-Net that can perform the mapping $F$.
\end{lemma}

\begin{theorem}
\label{thm:classical-PSBNN}
    (Theorem 1 by Sharma et al. \yrcite{do_BNN_need_fully_stochastic}). Let $X$ be a random variable taking values in $\mathcal{X}$, where $\mathcal{X}$ is a compact subspace of $\mathbb{R}^d$, and let $Y$ be a random variable taking values in $\mathcal{Y}$, where $\mathcal{Y} \subseteq \mathbb{R}^n$. Further, let $f_\theta: \mathbb{R}^m \times \mathcal{X} \rightarrow \mathcal{Y}$ represent a neural network architecture with a universal approximation property with deterministic parameters $\theta \in \Theta$, such that, for input $X=x$, the network produces outputs $f_\theta(Z, x)$, where $Z=\left\{Z_1, \ldots, Z_m\right\}, Z_i \in \mathbb{R}$, are the random variables in the network, which are Gaussian, independent of $X$, and have finite mean and variance.
    If there exists a continuous generator function, $\tilde{f}: \mathbb{R}^m \times$ $\mathcal{X} \rightarrow \mathcal{Y}$, for the conditional distribution $Y \mid X$, then $f_\theta$ can approximate $Y \mid X$ arbitrarily well. Formally, $\forall \varepsilon>0, \lambda<$ $\infty$,
    $$
    \begin{aligned}
    & \exists \theta \in \Theta, V \in \mathbb{R}^{m \times m}, u \in \mathbb{R}^m: \\
    & \sup _{x \in \mathcal{X}, \eta \in \mathbb{R}^m,\|\eta\| \leq \lambda}\left\|f_\theta(V \eta+u, x)-\tilde{f}(\eta, x)\right\|<\varepsilon .
    \end{aligned}
    $$
\end{theorem}
Now, we state the proof of our main theorem \ref{thm:main-theorem}:
\begin{proof}
Let \(\varepsilon > 0\). Let's assume that the endpoints of the SDE are \(t_1 = 0, t_2 = 0.5\), and let us define the decision regions $R_1=[0,0.5),R_2=[0.5,0.75),R_3=[0.75,1]$. Our proof explicitly constructs a set of parameters \(\theta, \sigma, \mathcal L\) which satisfy the desired property.\\
Before we start the proof, we assume that the input \(x\) is augmented to \(x_+ = [x^T,\mathbf{0}_{d_h-d_x}]^T = h_0 \in \R^{d_h}\), and \(w_t \in \mathbb R^{d_w}\), where \(d_h \geq d_x, d_w \geq m\) will be specified later. We also separate $\theta=[\theta_1,\theta_2,\theta_3]$ where we will choose $\theta_1,\theta_2,\theta_3$ later. Given a vector \(b \in \R^n\), we write \(b[i:j] := (b_i,b_{i+1}, \cdots, b_j) \in \R^{j - i + 1}\), and more generally, for a set of indices \(\mathcal I = \{ i_1 < i_2 < \cdots < i_k \} \subseteq \{1,\cdots,n\}\), we denote by \(b[\mathcal I] := (b_{i_1},b_{i_2},\cdots,b_{i_k}) \in \R^k\). We denote by \(\mathbf 0_{a\times b} \in \mathbb R^{a\times b}\) the \(a \times b\) matrix with 0 entries. Also, we denote by \(\mathbf 0_{n} \in \mathbb R^{n}\) the \(n \) dimension row vector with 0 entries.

We assume that the \(d_w\)-dimensional Brownian Motion \(B_t\) is generated independently from \(X\).

The proof is divided into three steps, corresponding to the three time regions \(R_1 = [0,0.5), R_2 = [0.5,0.75),R_3 = [0.75,1]\).

\textbf{Step 1}. We let $a(x)=[x^T,\mathbf{0}_{d_h-d_x}]^T$, which gives the initial values $h_0=[x^T,\mathbf{0}_{d_h-d_x}]^T$. We also set $w_0=\mathbf{0}_{d_w}^T$, and
\begin{equation}
\label{eq:vol_weights}
    g_p(t,w_t)=
    \begin{pmatrix}
        \sigma I_{m_{1}}\quad \mathbf{0}_{m_1\times m_2} \\
        \mathbf{0}_{m_2\times m_1}\quad \mathbf{0}_{m_{2}\times m_2}\\
    \end{pmatrix} =: \Sigma
\end{equation}
where \(\sigma > 0\), \(m_2 = d_w - m_1\) and $m_1$ will be specified later.
We choose \(\theta_1\) such that \(f_q (t,w_t;[\theta_1,\theta_2,\theta_3]) \equiv \mathbf{0}_{d_w}^T\) on \(R_1\), and we define 
\[
f_h(t,h_t;w_t) \equiv [\mathbf{0}_{d_x + r},w_t^T[1:m]]^T
\]
where \(r = d_h - d_x - d_w\) will be specified later. It is then easy to see that, letting the system evolve according to Equation \eqref{eq:system_sde}, we get at time \(t_2 = 0.5\)
\begin{align}
    w_{0.5} = \left[ \sigma B^T_{0.5}[1:m_1], \mathbf{0}_{m_2} \right]^T \quad
    h_{0.5} = \left[ x^T,\mathbf{0}_{r},\int_{0}^{0.5}\sigma B_t^T [1:m]dt\right ]^T \label{eq:proof-h05}
\end{align}

Since \(B_t\) is a centered Gaussian process with i.i.d coordinates, then \(\int_{0}^{0.5}\sigma B_t[1:m] dt\) is a Gaussian random vector with mean \(\mathbf 0 _{m}^T\) and covariance \(\sigma_0^2 I_{m}\). Therefore, taking \(\sigma = (\sigma_0)^{-1}\), we have 
\[
\eta := \int_0^{0.5} \sigma B_t[1:m] \mathrm d t \sim \mathcal N (\mathbf 0 _{m}^T, I_{m})
\]

\textbf{Step 2}. We now detail the construction on \(R_3 = [0.75,1]\). Note that in the interval \((0.5,1]\), $g_p \equiv 0$ which means $w_t$ is no longer stochastic given $w_{0.5}$. Assume that \(h_{0.75} = h_{0.5}\), which is trivially achieved by setting \(f_h(t,h_t;w_t) \equiv \mathbf{0}_{d_h}^T\) for $t\in R_2$. Notice that \(\eta = h_{0.75}[d_x+r + 1 : d_x + r + m]\) by \eqref{eq:proof-h05}, and recall that \(\eta \sim \mathcal{N}(\mathbf 0 _{m}^T,I_m)\). Using Theorem \ref{thm:classical-PSBNN} by Sharma et al. \yrcite{do_BNN_need_fully_stochastic}, we know that for any \(\lambda > 0\) there exists a neural network \(\hat{f}:\mathbb{R}^{d_x+m}\rightarrow\mathbb{R}^{d_y}\), which takes as input \(\eta\) and \(x\) such that
\begin{equation}
\label{eq:noise-approx}
    \sup _{x \in \mathcal{X}, \eta \in \mathbb{R}^m,\|\eta\| \leq \lambda}\left\|\hat{f}(\eta, x)-\tilde{f}(\eta, x)\right\|<\varepsilon/2 .
\end{equation}
We take \(\lambda\) large enough so that \(\mathbb P (||\eta|| \leq \lambda) \geq 1 - \varepsilon / 2\), and let \(A_1 = \{ ||\eta|| \leq \lambda \}\). 

Using lemma \ref{lemma:node-approx-net}, we can get hold of a linear-capped \((d+m+d_y)-\)dimensional Neural ODE that can reproduce \(\hat{f}\), which is given by
\vspace{-0.05in}
\begin{equation}
\label{eq:node}
    \mathrm d x_t=f_{\gamma}(t,x_t)\mathrm d t,\quad x_0=[x^T,\mathbf{0}_{d_y},\eta^T]^T 
\end{equation}

where \(f_\gamma : \mathbb R \times \mathbb R^{d+m+d_y} \rightarrow \mathbb R^{d + m + d_y}\) is a neural network with parameters \(\gamma \in \mathbb R^{m_\gamma}\), which is continuous with respect to its parameters. 
We are now able to specify the constants $m_1$, $m_2$, $r$, \(d_w\) and \(d_h\). We take
\[
m_1 = \max(m,m_\gamma),m_2=m_\gamma,r=d_y, d_w=m_1+m_2, d_h = d_x + m + r
\]
Now we construct $f_h$ when $t\in R_3$. Note that if we define $d_h=d_x+m+r$ we will have
$$
h_{0.75}=[x^T,\mathbf{0}_{d_y},\eta^T]^T
$$
Therefore, we can take
\begin{equation}
\label{eq:f_h in R3}
f_h(t,h_t;w_t)=f_h^1(t,h_t;w_t[1:m_{\gamma}])
\end{equation}
for $t\in R_3$ where \(f_h^1(t,h_t;w_t[1:m_{\gamma}])\) has the same dependency structure with respect to the first $m_\gamma$ entries of \(w_t\) as \(f_\gamma\) with respect to \(\gamma\), for \(t \in R_3\). That is,
$$
f_h^1(t,h_t;\gamma)=f_{\gamma}(t,h_t),t\in R_3
$$
We define the corresponding linear operator applied to the output of this augmented Neural ODE as \(\mathcal L\) which is the linear operator applied to the output of \eqref{eq:node}.


We set \(\theta_3\) such that \(f_q(t,w_t;[\theta_1,\theta_2,\theta_3]) \equiv \mathbf 0_{d_w}^T\) on \(R_3\), hence \(w_t\) remains constant on that region.
Therefore, as $m_1=\max(m,m_{\gamma})$, assuming that \(w_{0.75}[1:m_\gamma]\) is close enough to \(\gamma\), using the continuity of the output of the Neural ODE \eqref{eq:node} with respect to \(\gamma\), we eventually get that the output of our model satisfies 
\begin{equation}
    \forall \omega \in A_1, \forall x \in \mathcal X, \;||\mathcal L(\psi_1(\omega,x)) - \tilde{f}(\eta(\omega),x)|| \leq ||\mathcal L(\psi_1(\omega,x)) - \hat{f}(\eta(\omega),x)||  + ||\hat{f}(\eta(\omega),x) - \tilde{f}(\eta(\omega),x) || \leq \varepsilon
\end{equation}

All we are left to do is to show that, given \(\delta > 0\), we can choose \(\theta_2\) and \(A \in \mathcal F\) such that, \(\mathbb P (A) \geq 1 - \varepsilon\), and, for \(\omega \in A\)
\begin{equation}
\label{eq:obj-w075}
||\gamma - w_{0.75}[1:m_\gamma] || \leq \delta
\end{equation}

\vspace{-0.1in}
\textbf{Step 3}. We now detail the construction of $f_h$ and $f_q$ in the region \(R_2 = [0.5,0.75)\). 
Recall that we set \(f_h(t,h_t;w_t) \equiv \mathbf{0}_{d_h}^T \) on \(R_2\), all we need to do is defining $f_q$.
We shall consider three different cases:
\begin{itemize}
\vspace{-0.15in}
    \item[(i)]fixing $w_{0.5}$ \textit{a priori};
    \item[(ii)]taking $w_{0.5}$ as a fixed learned parameter.
    \item[(iii)] taking $w_{0.5}$ as the output of the SDE \eqref{eq:weights_posterior}, that is, taking $w_{0.5}=[\sigma B^T_{0.5}[1:m_1], \mathbf{0}_{m_2}]^T$.
\end{itemize}
 Note that we do not change $h_{0.5}$ accordingly, i.e., the hidden state is not being reset.
Let's discuss each case separately.

\textit{Case (i)}. Without loss of generality we can set $w_{0.5}\equiv \mathbf{0}^T_{d_w}$, then we can simply set \(\theta_2\) such that 
\[
\forall t \in R_2, \quad f_q(t,w_t;[\theta_1,\theta_2,\theta_3]) = \begin{pmatrix}
    4\gamma \\
    \mathbf 0^T_{d_w-m_{\gamma}}
\end{pmatrix}
\]
This leads to \(w_{0.75} = [\gamma^T, \mathbf 0^T_{d_w-m_{\gamma}}]^T\), which fulfills \eqref{eq:obj-w075} with \(A = A_1\).\\
\textit{Case (ii)}. If $w_{0.5}$ is learnable, we can simply take $w_{0.5}=[\gamma^T,\mathbf{0}_{d_w-m_{\gamma}}]^T$ and $f_q\equiv \mathbf{0}_{d_w}^T$ which leads to $w_{0.75}=[\gamma^T,\mathbf{0}_{d_w-m_{\gamma}}]^T$ and $A=A_1$.

\textit{Case (iii)}. Assume that \(w_{0.5}=[\sigma B^T_{0.5}[1:m_1], \mathbf{0}_{m_2}^T]^T\). As $m_1=\max(m,m_{\gamma})$, we can take \(\Lambda \geq 0\) large enough such that 
\[
\mathbb P \left( ||\sigma B_{0.5}[1:m_\gamma] || \leq \Lambda \right) \geq 1 - \frac{\varepsilon}{2}
\]
Let \(A_2 = \{ ||\sigma B_{0.5}[1:m_\gamma] || \leq \Lambda \}\). Let \(\phi: \mathbb R^{m_\gamma} \rightarrow \mathbb R^{m_\gamma}\) be any homeomorphism transformation mapping \(\mathcal B(0,\Lambda)\) onto \(\mathcal B (\gamma, \delta / 2)\) (for instance, \(\phi (y) = \gamma + \frac{\delta}{2\Lambda} y\)). Then, using lemma \ref{lemma:homeo-approx-net}, we can choose \(\theta_2\) such that the \(d_w-\)Neural ODE, integrated between \(t_2 = 0.5\) and \(t_3 = 0.75\)
\begin{align*}
    \frac{\mathrm d w_t}{\mathrm d t} = f_q(t,w_t;[\theta_1,\theta_2,\theta_3]) \qquad
    w_{0.5} = [\sigma B^T_{0.5}[1:m_1], \mathbf{0}_{m_\gamma}]^T
\end{align*}
maps, for any \(\omega \in A_2\), \(w_{0.5}\) to \(w_{0.75}\) such that \(w_{0.75}[1:m_\gamma] \in \mathcal B (\gamma, \delta/2)\), hence we have \eqref{eq:obj-w075} with \(A = A_1 \cap A_2\). 
\end{proof}

\section{More on Uncertainty Quantification}
\label{app: roc-curves}
\vspace{-0.1in}
\begin{figure}[h!]  
    \centering
    \includegraphics[width=0.24\textwidth]{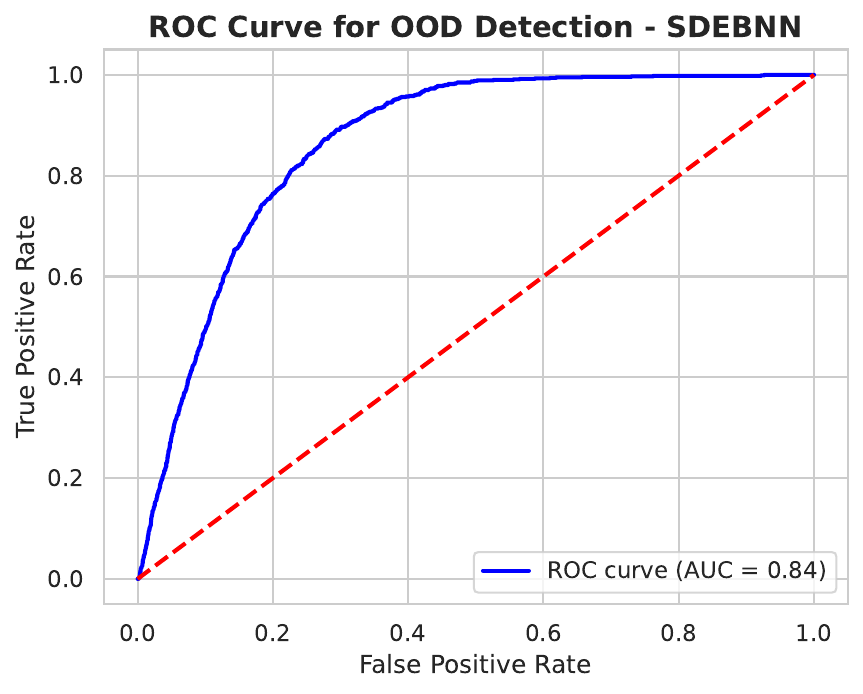}
    \includegraphics[width=0.24\textwidth]{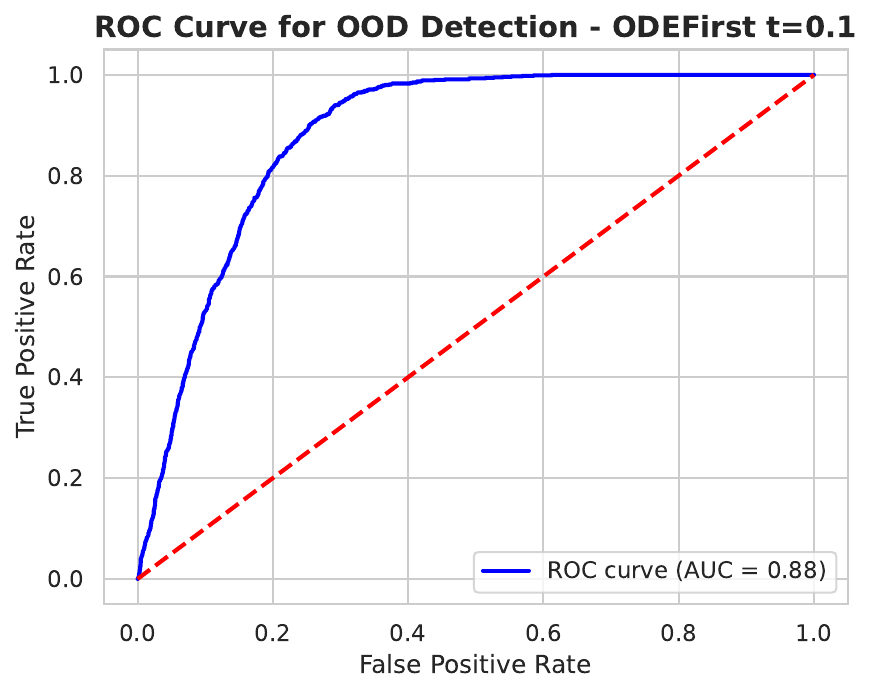}
    \includegraphics[width=0.24\textwidth]{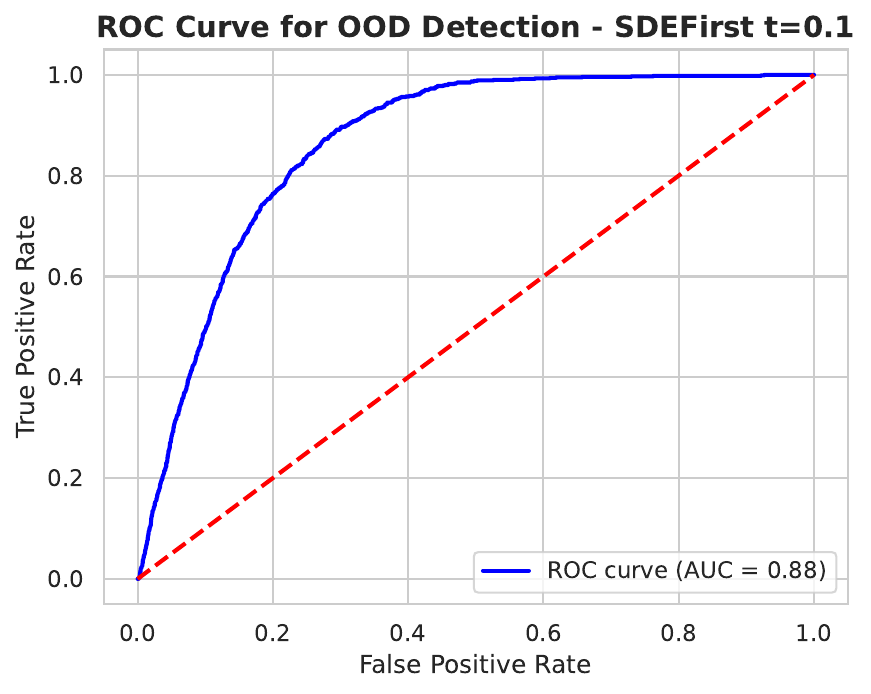}
    \includegraphics[width=0.255\textwidth]{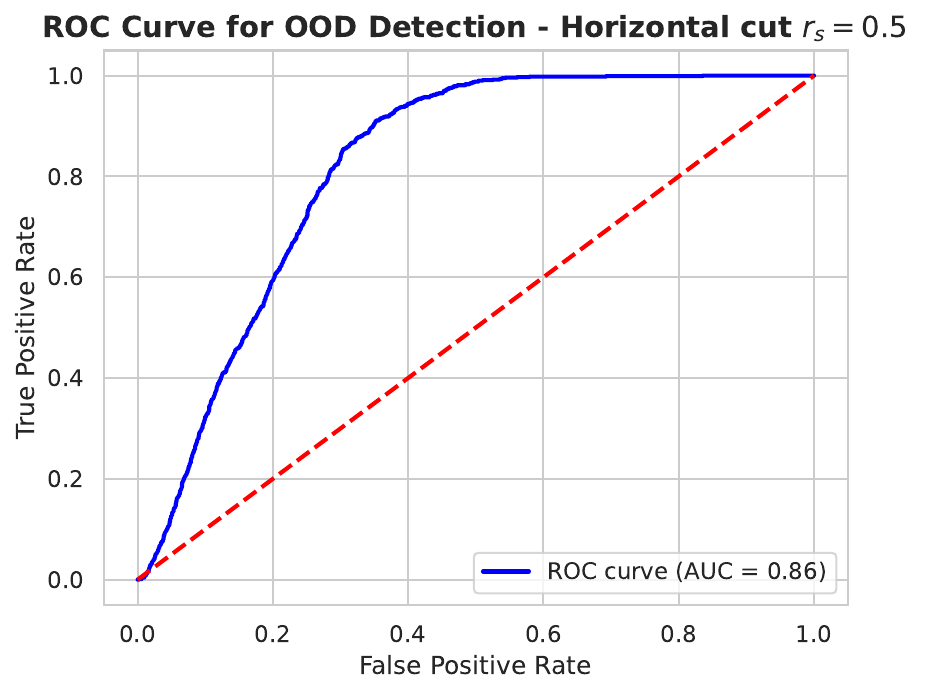}
    \vspace{-0.1in}
    \caption{ROC curves for OOD detection performance across three models: SDE-BNN, PSDE-BNN ODEFirst with \(r_s=0.1 \), PSDE-BNN SDEFirst with \(r_s=0.1\), and the PSDE-BNN with horizontal cut of the weights (\(r_s=0.5\)). The curves quantify each model's ability to differentiate between in-distribution (CIFAR-10 test set) and OOD samples, with the AUC metric reflecting the discrimination power. A higher AUC value indicates greater efficacy in distinguishing OOD from in-distribution data.}
    \label{fig: roc-curves}
\end{figure}

\newpage
\section{Further Training Visualisations}
\label{app: train-plots}

\begin{figure}[h!]  
    \centering
    \includegraphics[width=0.3\textwidth]{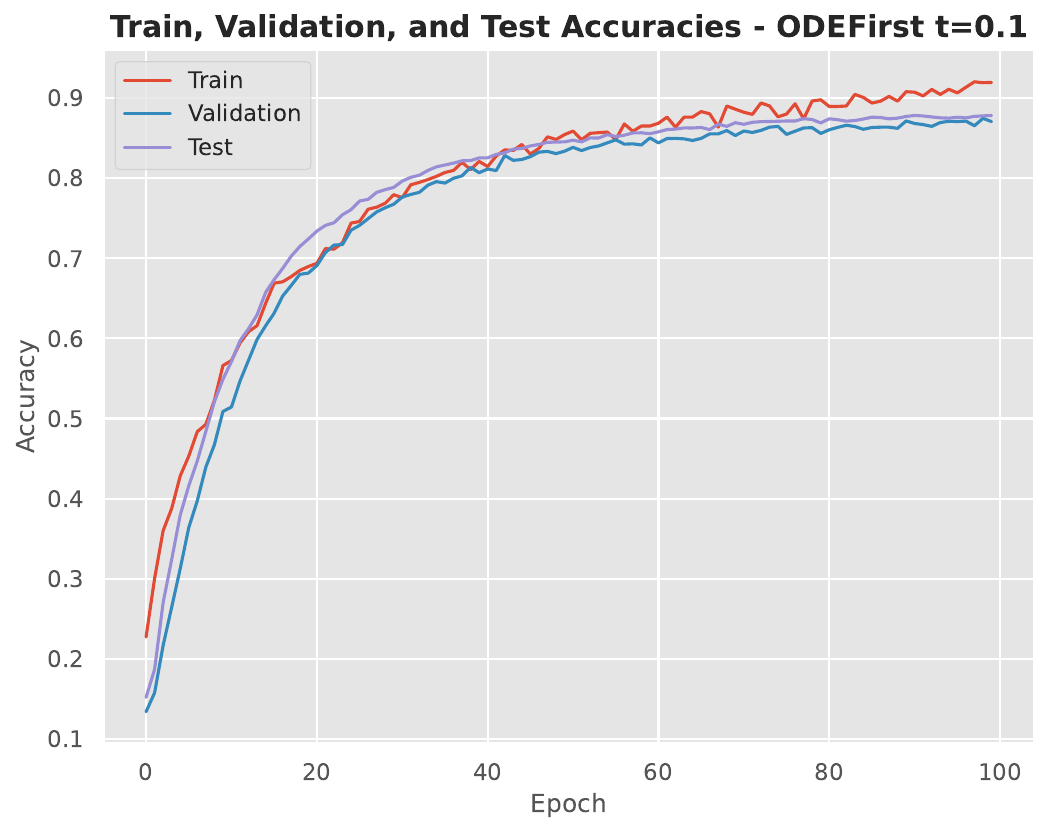}
    \hspace{0.02\linewidth} 
    \includegraphics[width=0.3\textwidth]{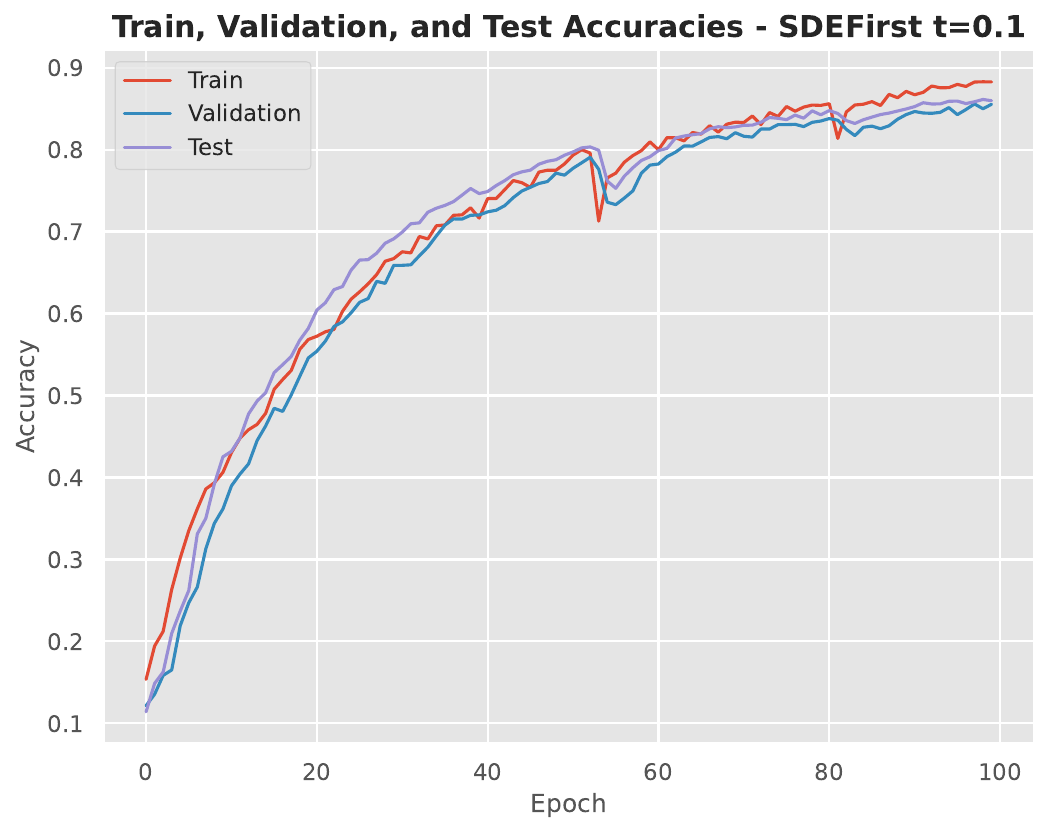}
    \hspace{0.02\linewidth} 
    \includegraphics[width=0.3\textwidth]{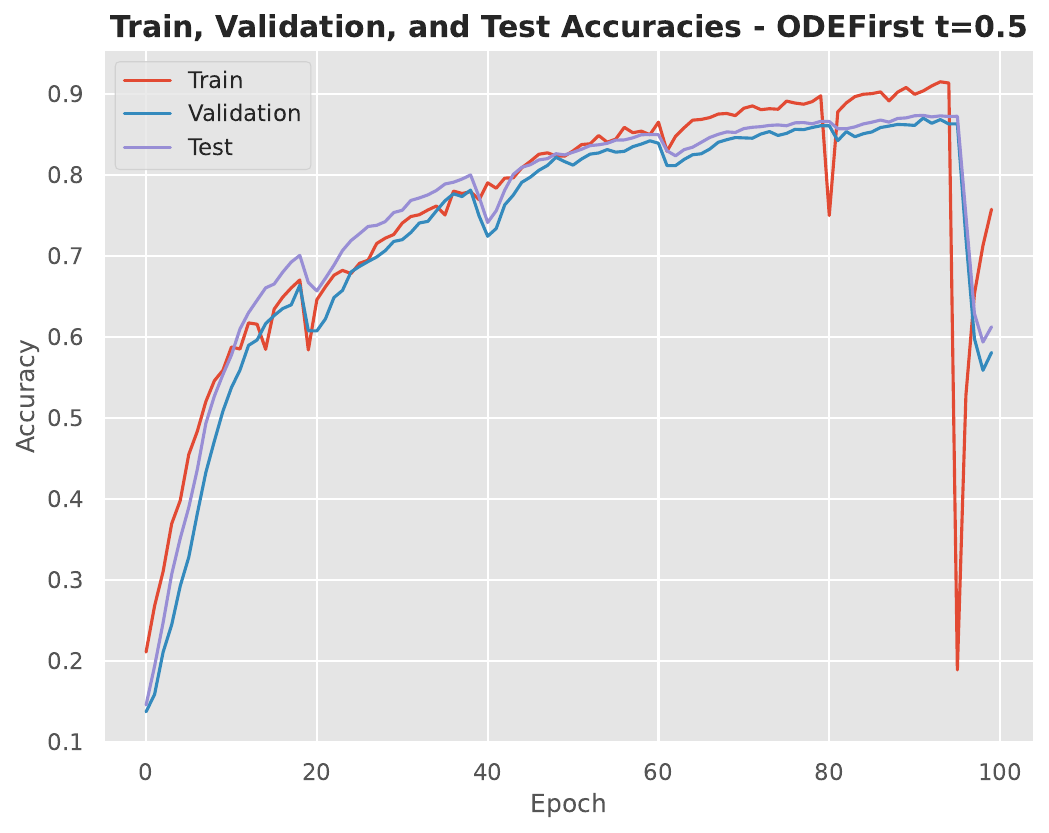}

    \includegraphics[width=0.3\textwidth]{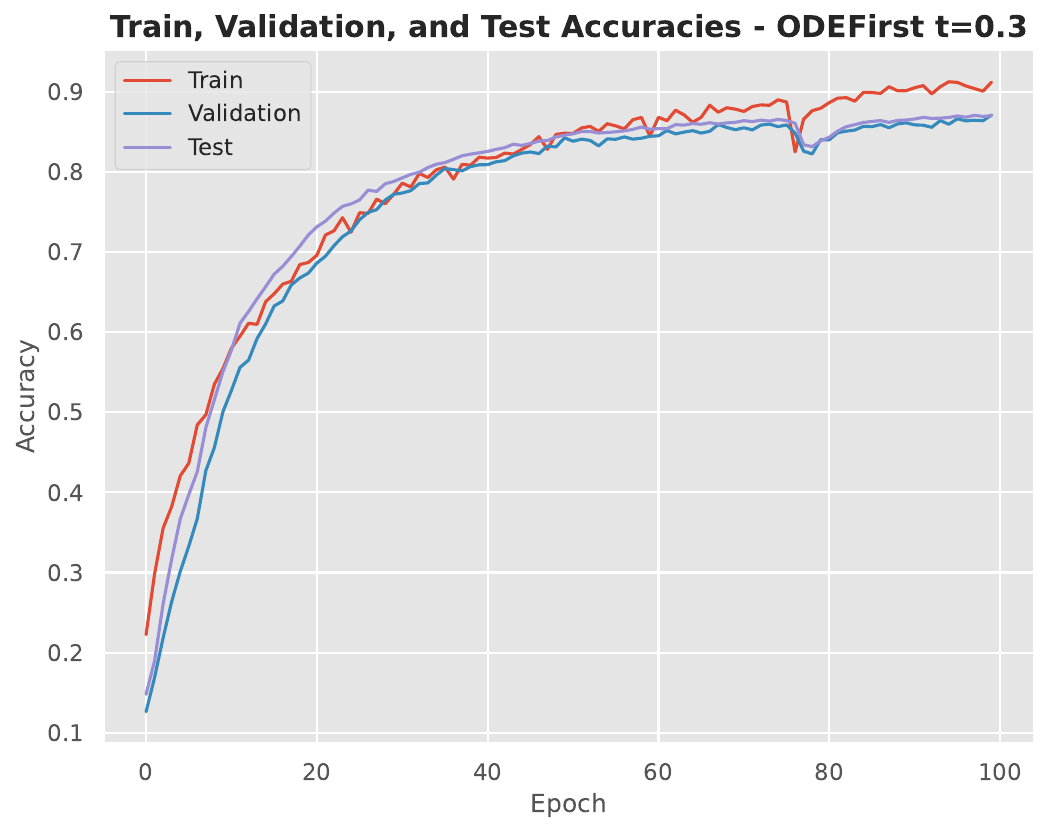}
    \hspace{0.02\linewidth} 
    \includegraphics[width=0.3\textwidth]{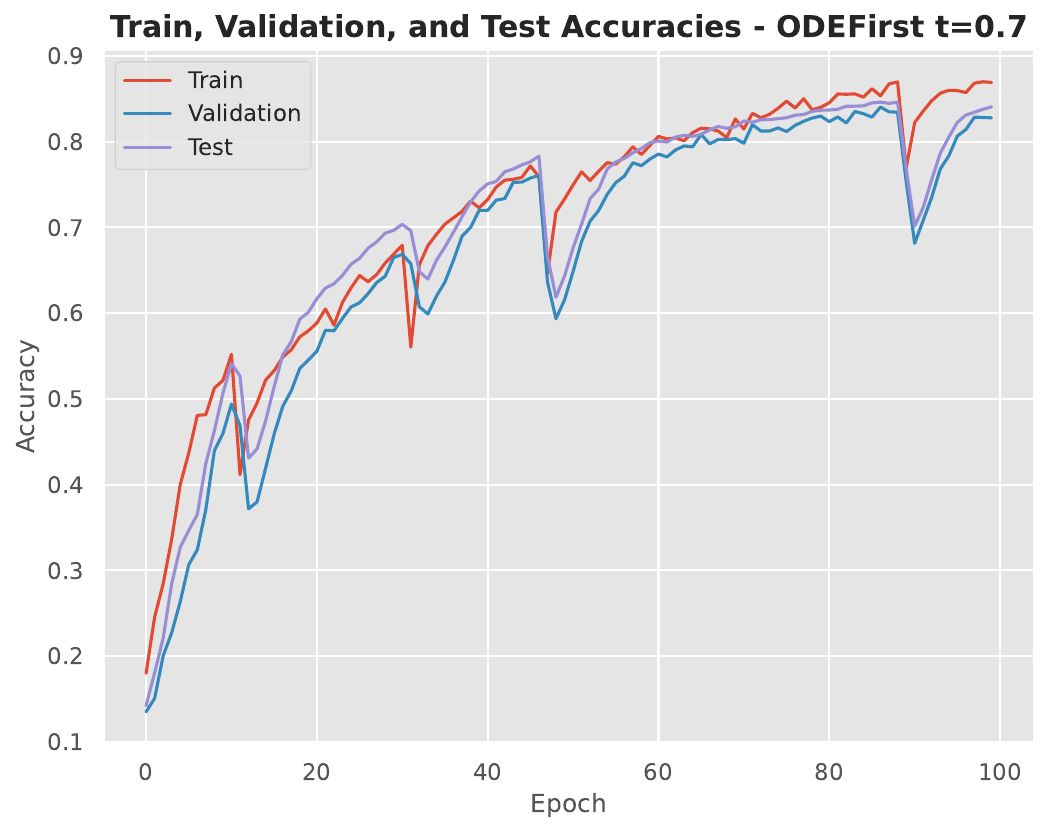}
    \hspace{0.02\linewidth} 
    \includegraphics[width=0.3\textwidth]{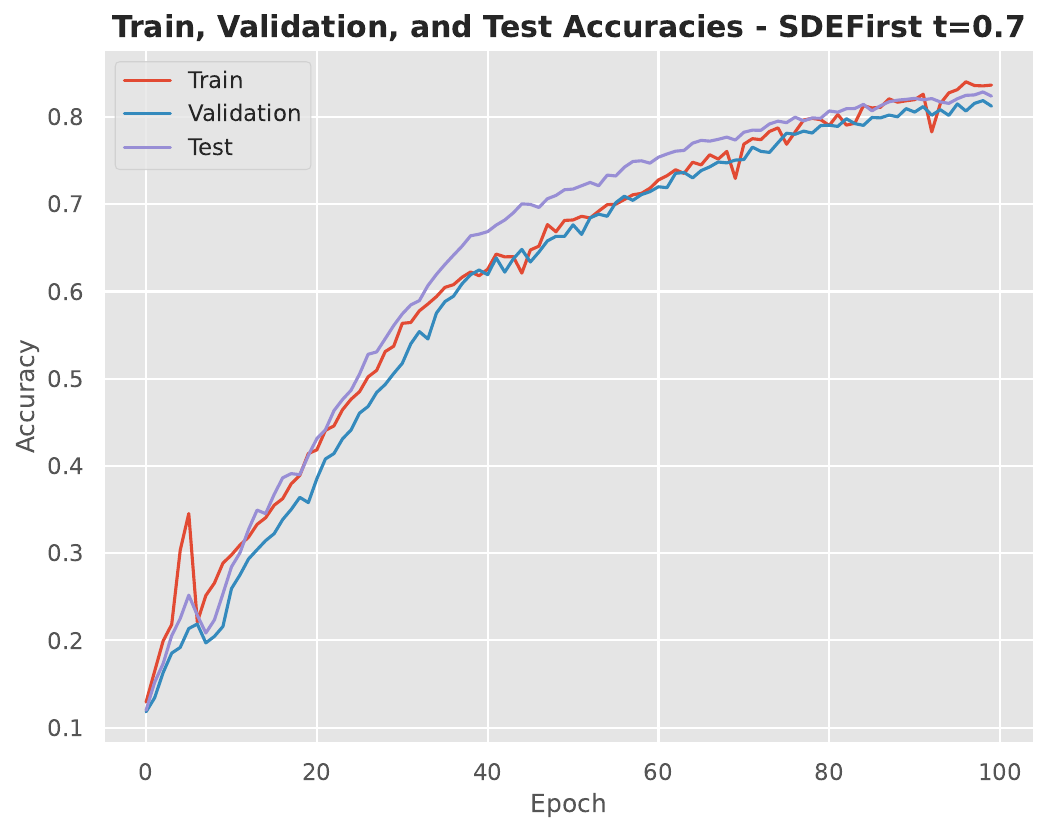}
    \caption{Training, test, and validation accuracy evolution over the first 100 training epochs on the CIFAR-10 dataset. Plots aim to illustrate the rate of learning and convergence stability across epochs and highlight how quickly each model achieves competitive performance. We can observe some symptoms of numerical instability which are inherent to differential equations numerical solvers. We overcome this issue by storing checkpoints only at the best validation accuracy.}
    \label{fig: acc-curves}
\end{figure}

\begin{figure*}[h!]  
    \centering
    \includegraphics[width=0.3\textwidth]{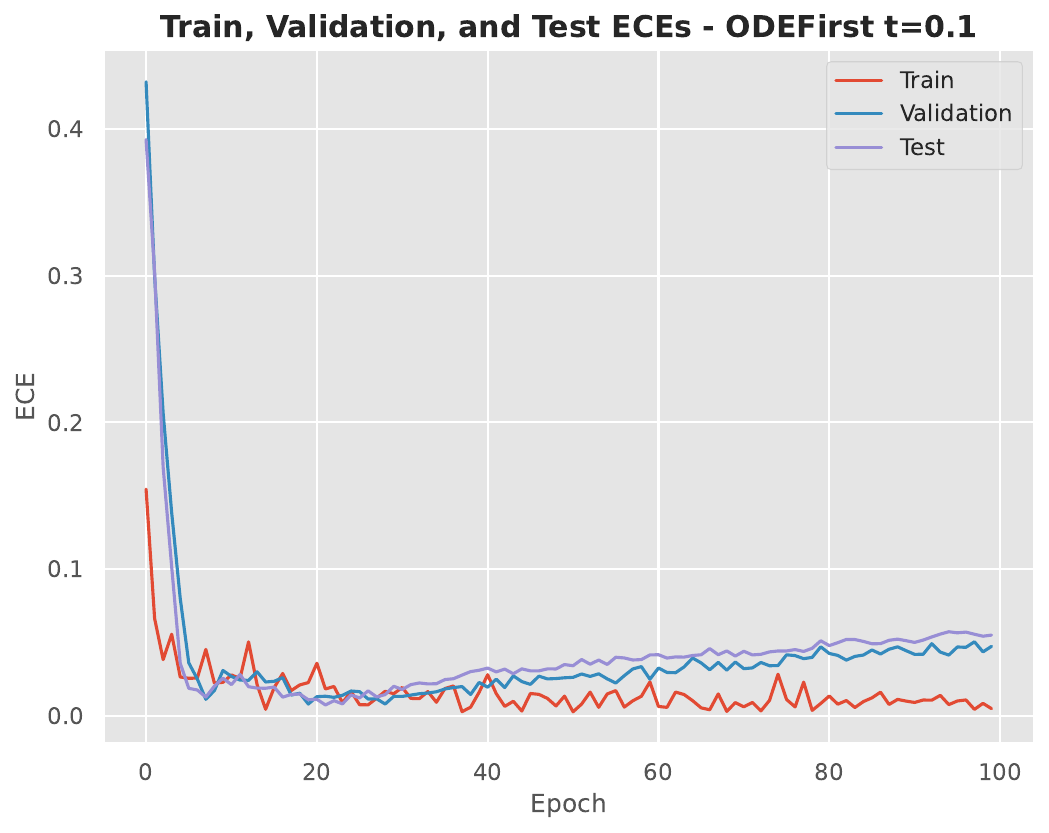}
    \hspace{0.02\linewidth} 
    \includegraphics[width=0.3\textwidth]{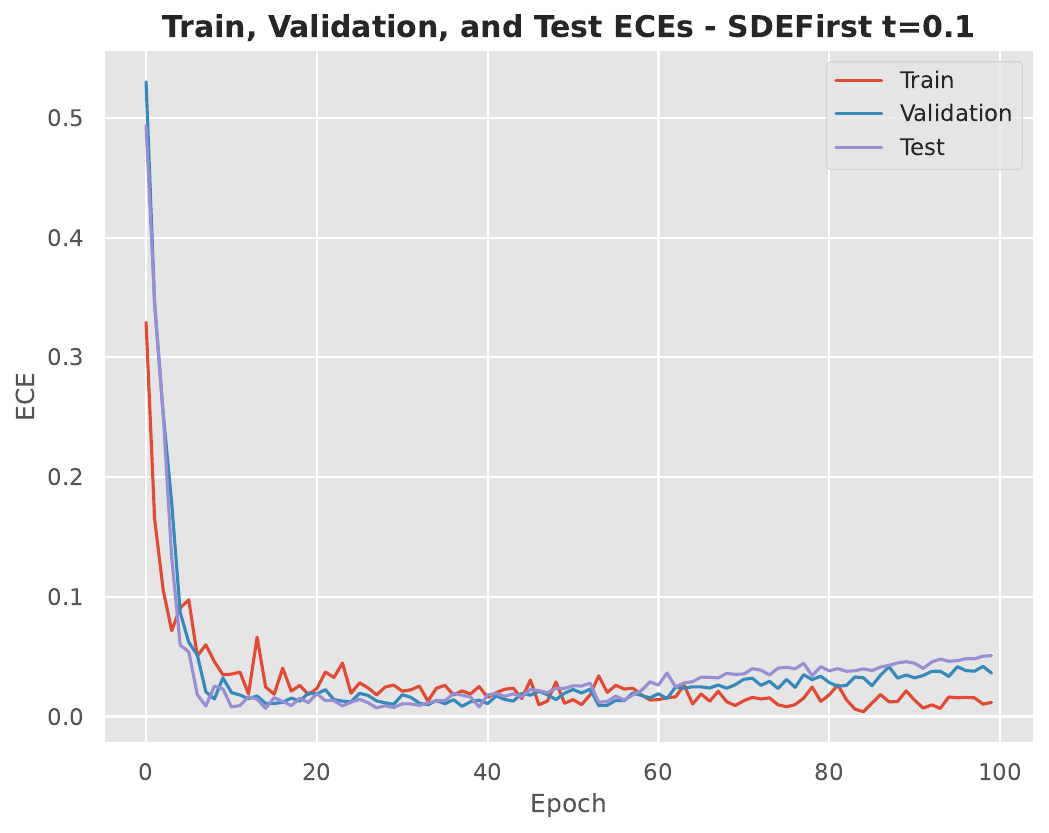}
    \hspace{0.02\linewidth} 
    \includegraphics[width=0.3\textwidth]{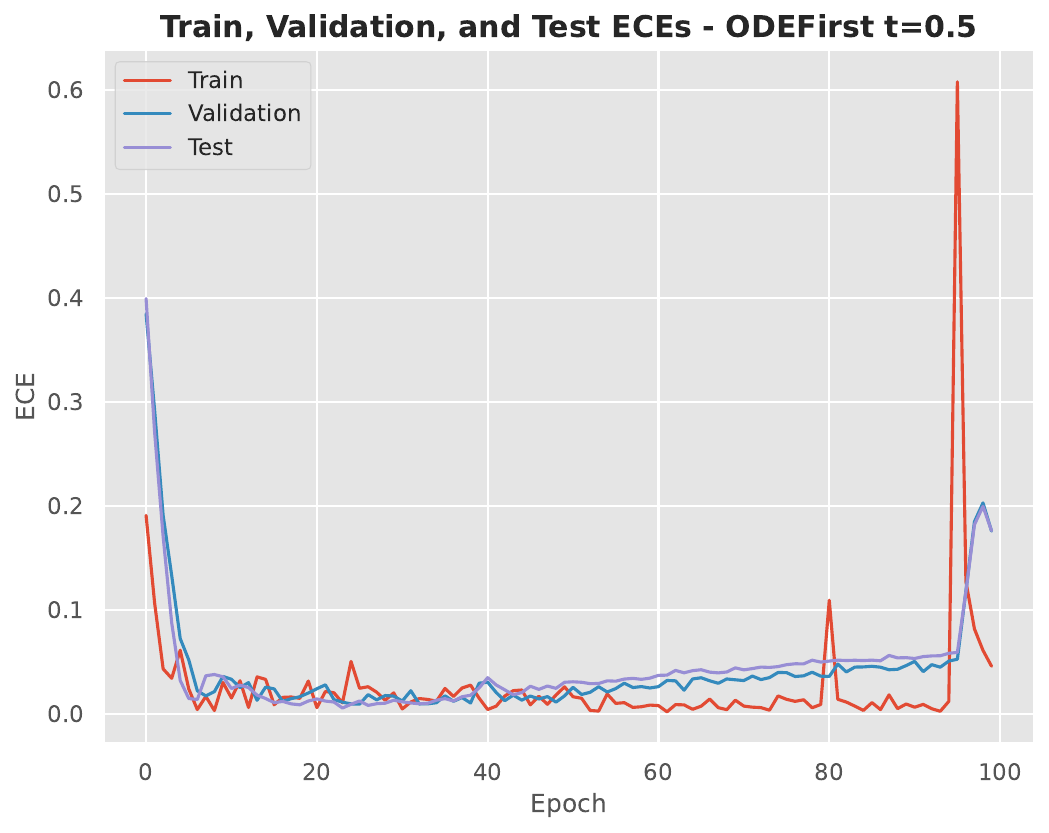}

    \includegraphics[width=0.3\textwidth]{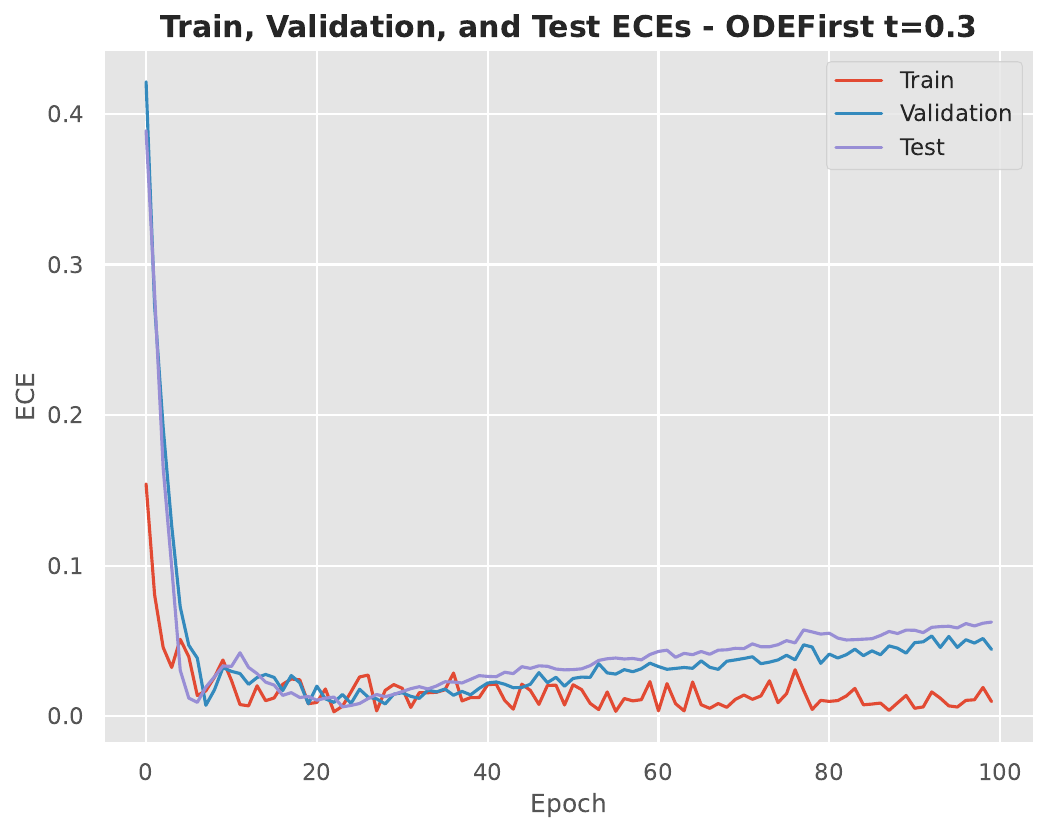}
    \hspace{0.02\linewidth} 
    \includegraphics[width=0.3\textwidth]{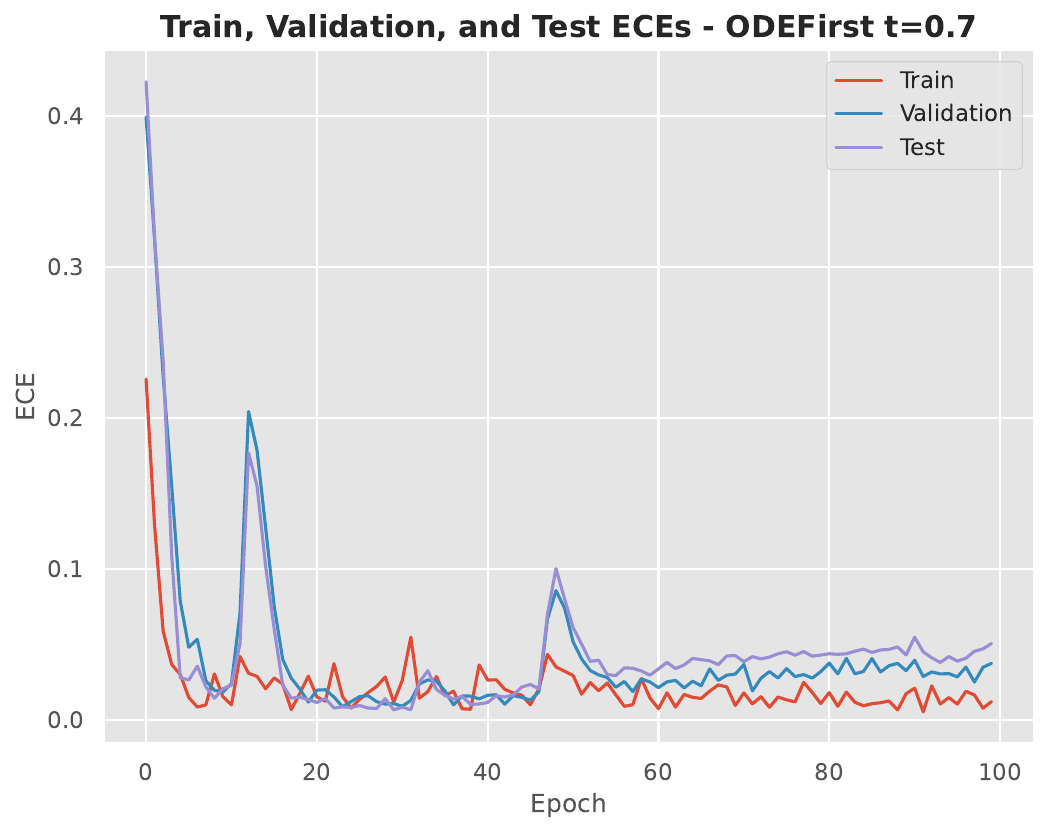}
    \hspace{0.02\linewidth} 
    \includegraphics[width=0.3\textwidth]{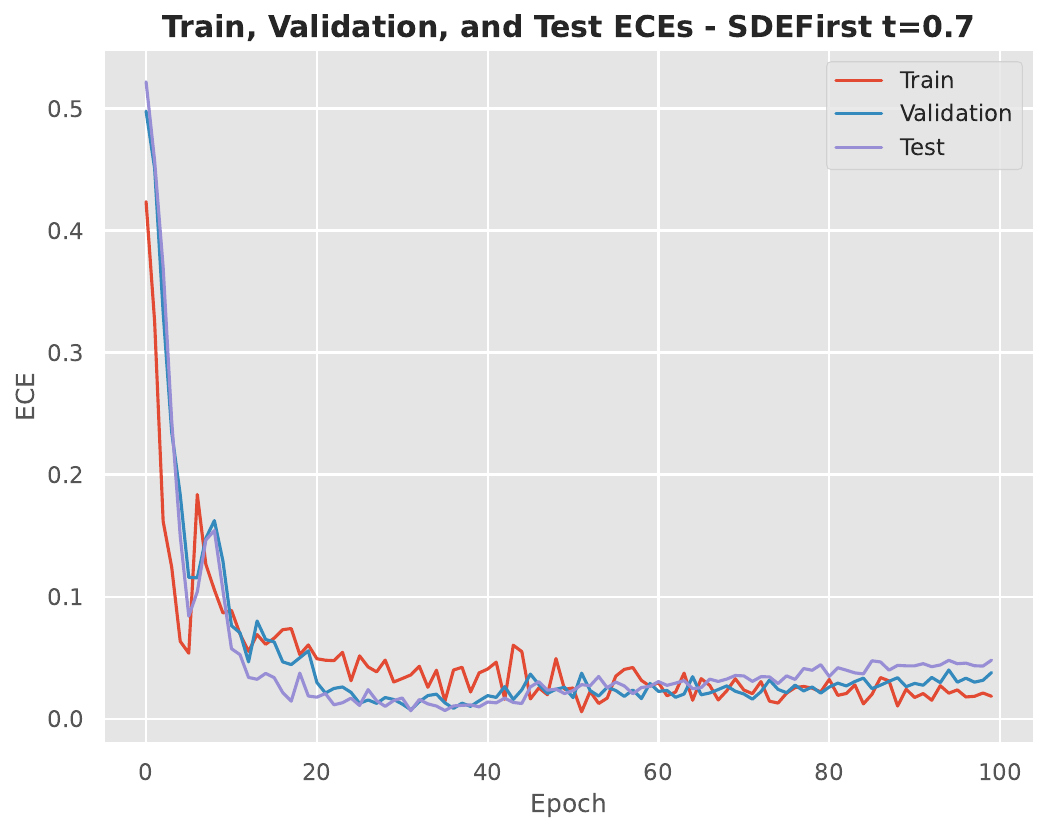}
    \caption{Training, test, and validation expected calibration error evolution over the first 100 training epochs on the CIFAR-10 dataset.}
    \label{fig: ece-curves}
\end{figure*}

\newpage
\section{Further Ablations}
\label{app: ablation}

\begin{table*}[ht]
\centering
\begin{tabular}{lccccc}
\hline
\textbf{Models} & \textbf{Accuracy (\%)} & \textbf{ECE $(\times 10^{-2})$} & \textbf{Pred. Entropy (OOD)} & \textbf{Epoch Time (s)} & \textbf{Inference Time (s)} \\
\hline
SDE-BNN & 86.04 \(\pm 0.25\) & 9.13 \(\pm 0.28\)  & 0.56 \(\pm\) 0.36 & 371.9 & 43.7 \\

ODEFirst (10\%) & \textbf{87.84} \(\pm\) \textbf{0.08} & 4.73 \(\pm\) 0.07 & 0.79 \(\pm\) 0.37 & \textbf{289.4} & \textbf{31.7} \\

ODEFirst (30\%) & 87.10 \(\pm\) 0.13 & 4.43 \(\pm\) 0.09 & 0.67 \(\pm\) 0.35 & 310.3 & 35.4 \\

ODEFirst (50\%) & 87.23 \(\pm\) 0.11 & 5.06 \(\pm\) 0.16 & 0.67 \(\pm\) 0.33 & 331.9 & 38.9 \\

ODEFirst (70\%) & 84.47 \(\pm\) 0.15 & 3.58 \(\pm\) 0.14 & 0.82 \(\pm\) 0.39 & 353.2 & 42.7 \\

SDEFirst (10\%) & 85.34 \(\pm 0.21\) & 3.56 \(\pm\) 0.15 & \textbf{0.95} \(\pm\) \textbf{0.38} & 295.1 & \textbf{31.7} \\

SDEFirst (30\%) & 87.19 \(\pm 0.24\) & 4.09 \(\pm\) 0.17 & 0.69 \(\pm\) 0.35 & 314.5 & 35.5 \\

SDEFirst (50\%) & 82.24 \(\pm 0.32\) & \textbf{2.24} \(\pm\) \textbf{0.17} & 0.73 \(\pm\) 0.34 & 334.9 & 39.2 \\

SDEFirst (70\%) & 83.00 \(\pm 0.17\) & 3.77 \(\pm\) 0.22 & 0.87 \(\pm\) 0.32 & 413.1 & 42.7 \\
\hline
\end{tabular}
\vspace{-0.15in}
\caption{Comparison of CIFAR-10 model performance across various metrics and averaged across 3 different seeds. The KL coefficient is set to be a function of the stochasticity ratio to be proportional to the length of the SDE.}
\label{tab: ablation}
\vspace{-0.15in}
\end{table*}

\subsection{Computational Complexity Overview}\label{app: comp-complex}

We will now provide a breakdown of the time complexities for each model. Let \(D_h, D_w\) denote the dimension of the hidden state \(h_t\) and weights \(w_t\) respectively, and let \(D = D_h + D_w\). At inference, we have:

\begin{itemize}
  \item \textbf{Shared Complexity in deterministic part (i.e. induced by \(\Delta t\) in equation \eqref{eq:system_sde})}: Both our and \cite{xu2022infinitely}'s approaches share the complexity \(\mathcal{O}(F(D)T)\) for the deterministic part of the Euler-Maruyama discretization, where \(T\) is the total number of time steps and \(F(D)\) represents the cost of evaluating the functions \(f_g\) and \(f_h\) at each time step.
  \item \textbf{SDE Part Complexity (i.e. induced by \(dB_t\) in equation \eqref{eq:system_sde}) - Original SDE-BNN Model}: The SDE-BNN introduces additional complexity proportional to \(D_wT\), corresponding to an additional \(D_w\)-dimensional Gaussian random variable to simulate at each time step (stochastic part of the Euler-Maruyama discretization)
  \item \textbf{SDE Part Complexity - PSDE-BNN Model}
    \begin{itemize}
      \item \textbf{Vertical Cut}: Introducing vertical cut partitions the time into deterministic and stochastic phases, reducing the SDE part complexity to \(\mathcal{O}((D_wT_{\text{subset}})\) where \(T_{\text{subset}}\) is the stochastic phase duration. We can write this as \(\mathcal{O}(rD_wT)\), given a stochastic ratio \(r = \frac{T_{\text{subset}}}{T} \leq 1\).
      \item \textbf{Horizontal Cut}: This model limits the stochastic dimensions to \(D_s < D_w\), reducing the SDE part complexity to \(\mathcal{O}(D_sT)\) (or \(\mathcal{O}(rD_sT)\) if combined with a vertical cut).
    \end{itemize}
\end{itemize}

We can then summarize the total time complexities in the following way:
\begin{table}[h]
\centering
\begin{tabular}{ll}
\toprule
\textbf{Model}                          & \textbf{Time Complexity}                \\ \midrule
SDE-BNN                                 & \(\mathcal{O}((F(D) + D_w)T)\)         \\ 
PSDE-BNN with Vertical cut              & \(\mathcal{O}((F(D) + D_w)T)\)         \\ 
PSDE-BNN with Horizontal cut            & \(\mathcal{O}((F(D) + D_s)T)\)         \\ 
PSDE-BNN with Vertical and Horizontal cuts & \(\mathcal{O}((F(D) + rD_s)T)\)     \\ \bottomrule
\end{tabular}
\end{table}

where \(r < 1\) and \(D_s < D_w\).

The distinction between \(D_w\) and \(D_h\) is key, as \(D_h\) is typically much larger, making stochastic simulations notably more demanding. Thus, by reducing the stochastic portion, our models significantly enhance efficiency without compromising the depth of stochastic analysis.

\newpage
\newpage
\section{Experimental setting}
\label{app: experiments}

\begin{table*}[h!]
\centering
\caption{Hyperparameter settings for model evaluations in the classification tasks as presented in Table 1. Experiments were conducted on a single Nvidia RTX 3090 GPU within our compute clusters. Throughout the training, no scheduling was applied to the hyperparameters. Configurations that share similarities with other models are denoted by \texttt{<model>} notation, where additional parameters are specified.}
\label{tab: hyperparams}
\vspace{0.1in}
\begin{tabular}{@{}llll@{}}
\toprule
\textbf{Model} & \textbf{Hyper-parameter}  & \textbf{MNIST} & \textbf{CIFAR-10} \\

\cmidrule{1-2} \cmidrule(lr){3-4}
\textbf{SDE-BNN} & Learning Rate & 1e-3 & 7e-4 \\
        & Batch Size & 128 & 128 \\
        & Activation & softplus & softplus \\
        & Epochs & 100 & 300 \\
        & Augment dim. & 0 & 0 \\
        & KL coef. & 1e-3 & 1e-4 \\
        & \# Solver Steps & 60 & 30 \\
        & \# blocks & 1 & 2-2-2 \\
        & Drift $f_a$ dim. & 32 & 64 \\
        & Drift $f_w$ dim. & 1-32-1 & 2-128-2 \\
        & Diffusion $\sigma$ & 0.2 & 0.1 \\
        & \# Posterior Samples \hspace{0.2in} & 1 & 1 \\
\cmidrule{1-2} \cmidrule(lr){3-4}
\textbf{PSDE-BNN ODEFirst} & $<$SDE BNN$>$ & $<$SDE BNN$>$ & $<$SDE BNN$>$ \\
                  & Epochs & 30 & 100 \\
                  & Stochasticity Ratio ($r_s$) & 0.1 & 0.1 \\
                  & KL coef. & $\frac{1}{r_s}10^{-3}$ & $\frac{1}{r_s}10^{-4}$ \\
\cmidrule{1-2} \cmidrule(lr){3-4}
\textbf{PSDE-BNN SDEFirst} & $<$SDE BNN$>$ & $<$SDE BNN$>$ & $<$SDE BNN$>$ \\
                  & Epochs & 30 & 100 \\
                  & Stochasticity Ratio & 0.1 & 0.1 \\
                  & KL coef. & $\frac{1}{r_s}10^{-3}$ & $\frac{1}{r_s}10^{-4}$ \\
\cmidrule{1-2} \cmidrule(lr){3-4}
\textbf{PSDE-BNN SDEFirst} `\texttt{fix\_w}$_2$' & $<$SDE BNN$>$ & $<$SDE BNN$>$ & $<$SDE BNN$>$ \\
                                        & Epochs & 30 & 100 \\
                                        & Stochasticity Ratio & 0.1 & 0.1 \\
                                        & KL coef. & $\frac{1}{r_s}10^{-3}$ & $\frac{1}{r_s}10^{-4}$ \\
\cmidrule{1-2} \cmidrule(lr){3-4}
\textbf{PSDE-BNN Hor. Cut} & $<$SDE BNN$>$ & $<$SDE BNN$>$ & $<$SDE BNN$>$ \\
                                        & Epochs & 30 & 100 \\
                                        & Stochasticity Ratio & 0.5 & 0.5 \\
                                        & KL coef. & $\frac{1}{r_s}10^{-3}$ & $\frac{1}{r_s}10^{-4}$ \\
                                        & Drift $f_{w_{S/D}}$ dim. & 1-16-1 & 1-64-1 \\
\bottomrule
\end{tabular}
\end{table*}


\end{document}